\documentclass[journal]{IEEEtran}
\ifCLASSINFOpdf
\else
\fi

\usepackage{graphicx}
\usepackage{amsmath}
\usepackage{amssymb}

\usepackage{caption}
\usepackage{makecell}
\usepackage{booktabs}
\usepackage{subfig}
\usepackage{multirow}

\usepackage{multicol}
\usepackage{enumitem}

\usepackage{url}

\usepackage{breakurl}
\usepackage[breaklinks]{hyperref}

\usepackage{algorithmic}
\usepackage{algorithm}

\usepackage{adjustbox} 

\usepackage[table,x11names,dvipsnames,table]{xcolor}
\usepackage{tabu}

\begin{document}
%
\title{A Comprehensive Study on Face Recognition Biases Beyond Demographics}

%
%
%

\author{Philipp~Terh\"orst,
        Jan~Niklas~Kolf,
        Marco~Huber, 
        Florian~Kirchbuchner,
        Naser~Damer,
        Aythami Morales,
        Julian Fierrez,
        and~Arjan~Kuijper
\thanks{P. Terhörst, J. Kolf, M.Huber, F. Kirchbuchner, N. Damer, and A. Kuijper are with the Fraunhofer Institute for Computer Graphics Research IGD, Darmstadt, Germany and with the Technical University of Darmstadt, Darmstadt, Germany.
A. Morales and J. Fierrez are with Biometrics and Data Pattern Analytics Lab - BiDA Lab, Universidad Autonoma de Madrid, Madrid, Spain.
e-mail: \{philipp.terhoerst@igd.fraunhofer.de\}}}

%
%

\markboth{Journal of \LaTeX\ Class Files,~Vol.~14, No.~8, August~2015}%
{Shell \MakeLowercase{\textit{et al.}}: Bare Demo of IEEEtran.cls for IEEE Journals}
%



\maketitle

\begin{abstract}
Face recognition (FR) systems have a growing effect on critical decision-making processes. 
Recent works have shown that FR solutions show strong performance differences based on the user’s demographics.
However, to enable a trustworthy FR technology, it is essential to know the influence of an extended range of facial attributes on FR beyond demographics. 
Therefore, in this work, we analyse FR bias over a wide range of attributes. 
We investigate the influence of 47 attributes on the verification performance of two popular FR models. 
The experiments were performed on the publicly available MAADFace attribute database with over 120M high-quality attribute annotations. 
To prevent misleading statements about biased performances, we introduced control group based validity values to decide if unbalanced test data causes the performance differences. 
The results demonstrate that also many non-demographic attributes strongly affect the recognition performance, such as accessories, hair-styles and colors, face shapes, or facial anomalies. 
The observations of this work show the strong need for further advances in making FR system more robust, explainable, and fair.
Moreover, our findings might help to a better understanding of how FR networks work, to enhance the robustness of these networks, and to develop more generalized bias-mitigating face recognition solutions.

\end{abstract}

\begin{IEEEkeywords}
Face recognition, Biometrics, Bias, Fairness, Performance differences, Bias estimation, Soft-Biometrics
\end{IEEEkeywords}

%
\IEEEpeerreviewmaketitle

\vspace{-3mm}
\section{Introduction}












Large-scale face recognition systems are spreading worldwide \cite{DBLP:conf/icb/DamerT0K17}.
These systems have a growing effect on the daily life \cite{DBLP:journals/corr/abs-1804-06655} and are increasingly involved in critical decision-making processes, such as in forensics and law enforcement.
However, recent works \cite{faceAccurate,DBLP:journals/corr/abs-1809-02169,Furl2002FaceRA,Phillips:2011:OEF:1870076.1870082,pmlr-v81-buolamwini18a, garvie2016perpetual, richa} showed that current face recognition solutions possess biases leading to discriminatory performance differences \cite{Discrimination} based on the user's demographics \cite{Tan, Nixon}.

From a legal perspective, there are several regulations to prevent such discrimination, for instance, Article 7 of the Universal Declaration on Human Rights, Article 14 of the European Convention of Human Rights, or Article 71 of the General Data Protection Regulation (GDPR) \cite{Voigt:2017:EGD:3152676}.
Driven by (a) these legal efforts to guarantee fairness and (b) the findings that the performance of current face recognition solutions depends on the user's demographics, several approaches were proposed to mitigate demographics-bias in face recognition technologies.
This was achieved through adversarial learning \cite{gong2019debface,Liang_2019_CVPR, SensitiveNets, PrivacyNet}, margin-based approaches \cite{wang2019mitigate,DBLP:journals/corr/abs-1806-00194}, data augmentation \cite{WANG201935,Kortylewski_2019_CVPR_Workshops,DBLP:conf/cvpr/00010S0C19}, metric-learning \cite{DBLP:conf/iwbf/TerhorstD0K20}, or score normalization \cite{DBLP:journals/corr/abs-2002-03592}.
However, the research focus on demographic-bias does only tackle a minor proportion of all possible discriminatory effects.
Knowing the influence of an extended set of facial attributes on the face recognition performance will enable the development of accurate and less discriminatory face recognition systems.

In this work, we aim at investigating the face recognition bias based on a wide range of attributes beyond demographics. 
These biases might affect the fairness \cite{Discrimination} but also the security of face recognition systems \cite{marcel}. 
Biases can be identified as learning weakness to be exploited by users with malicious intentions (i.e. vulnerability attacks \cite{face-emo}). 
To be precise, we analyse the differential outcome as defined by Howard et al. \cite{DBLP:conf/btas/HowardSV19} of two popular face recognition models (FaceNet \cite{DBLP:journals/corr/SchroffKP15} and ArcFace \cite{Deng_2019_CVPR}) with regard to 47 attributes.
The experiments are conducted on the recently published and publicly available MAAD-Face\footnote{\url{https://github.com/pterhoer/MAAD-Face}} annotation database \cite{maadface} based on VGGFace2 \cite{Cao18}. 
It consists of over 120M high-quality attribute annotations for 3.3M face images.
For the experiments, several decision thresholds are taken into account to cover a wide range of applications.
To prevent misinterpretations of the results origin from testing data with (a) unbalanced label distributions or (b) attribute correlations, we (a) introduce control groups to derive a validity value for the recognition performance in the presence of a specific attribute and (b) analyse the pairwise correlations of the attribute annotations.
While (a) allows us to quantify results that arise from unbalanced testing data and prevent falsified statements about the attribute-related bias, (b) emphasize if an attribute bias might originate from a different (correlated) attribute.
Besides a detailed analysis, we present a visual summary that states the performance difference between samples with and without a specific attribute over the validity of the results. 
This aims to present the results in a compact and simply understandable manner.

The results support the findings of previous works stating that face recognition systems have to deal with demographic-biases \cite{DBLP:journals/corr/abs-2004-11246}. 
We differentiate between explicit demographic attributes such as gender, age or ethnicity and non-explicit demographic attributes such as accessories, hairstyles and -colors, face shapes, or facial anomalies. 
However, we have to consider that some of the non-explicit demographic attributes might be affected by implicit demographic covariates. 
For example, the hairstyle is highly affected by gender or ethnicity. 
The results demonstrate that also many of the non-demographic attributes strongly affect the recognition performance.
Investigating two face recognition models that differ only in the loss function used during training, we showed the effect of the underlying training principles on recognition.
While the triplet-loss based FaceNet model showed attribute-related differential outcomes that are relatively constant on several decision thresholds, the angular margin based ArcFace model showed differential outcomes that are often dependent on the used decision threshold.
Many performance differences affected by attributes could be explained through the attribute's relation to the visibility of a face, the temporal variability, and the degree of abnormality.
However, our experiment also reveals many unconventional results that future work have to address.
Our findings strongly motivate further advances in making recognition systems more robust against covariates \cite{DBLP:journals/tbbis/LuCCC19, faceWild}, explainable \cite{XAI, XAI-symbolic}, and fair \cite{DBLP:journals/corr/abs-2004-11246, bias-multi}.
We hope that these findings help to develop robust and bias-mitigating face recognition solutions and help also to move forward bias-aware and bias-mitigating technology in other AI application areas.

\vspace{-3mm}
\section{Related Work}

The phenomena of bias in face biometrics were found in several disciplines such as presentation attack detection \cite{DBLP:journals/corr/abs-2003-03151, intro-pad}, the estimation of facial characteristics \cite{BTAS_terhoerst, DBLP:conf/eccv/DasDB18}, and the assessment of face image quality \cite{Terhorst2020FaceQE}.
In general, one of the main reasons for bias might be the induction of non-equally distributed classes in training data \cite{Kortylewski_2019_CVPR_Workshops, DBLP:journals/corr/abs-1806-00194, InsideBias} that leads to differences in the recognition performance and thus, might have an unfair impact, e.g. on specific subgroups of the population.
Howard et al. \cite{DBLP:conf/btas/HowardSV19} introduced the terms differential performance and outcome for classifying biometric performance differentials that separately considers the effect of false positive and false negative outcomes.
They show that the often-cited evidence regarding biometric equitability has focused primarily on false-negatives.

Previous works on bias in face recognition \cite{9086771} mainly focused on the influence of demographics.
However, Terh\"orst et al. \cite{IJCB2020} demonstrated recently that more (non-demographic) characteristics are stored in face templates that might have an impact on the face recognition performance.
In the following, we will shortly discuss related works on estimating and mitigating bias in face biometrics.
For a more complete overview, we refer to \cite{9086771}.

\vspace{-2mm}
\subsection{Estimating Bias in Face Recognition}
\label{sec:RelatedWorkEstimatingBias}

In recent years, several works have been published that demonstrated the influence of demographics on commercial and open-sources face recognition algorithms.
Studies \cite{DBLP:conf/biosig/RicanekBS15, DBLP:conf/icb/MichalskiYM18, DBLP:conf/icb/DebN018, DBLP:conf/cvpr/SrinivasRMBK19} analysing the impact of age demonstrated a lower biometric performance on faces of children.
Studies \cite{DBLP:conf/cvpr/Vera-RodriguezB19, DBLP:journals/corr/abs-2002-02934, DBLP:journals/corr/abs-2008-06989, InsideBias} analysing the effect of gender on face recognition showed that the recognition performance of females is weaker than the performance on male faces.
Experiments without unbalanced data distributions and with an unbalanced towards female faces resulted in similar results \cite{DBLP:journals/corr/abs-2002-02934}.
In \cite{DBLP:journals/corr/abs-2008-06989}, experiments with a PCA-decomposition showed that females faces are intrinsically more similar than male ones.
Research analysing the impact of the user's ethnicity showed faces of ethnicities which were under-represented in the training process perform significantly weaker.
The same was found for darker-skinned cohorts in general \cite{9001031}.

More recent studies \cite{DBLP:journals/ivc/GeorgopoulosPP20, 6327355, 8638319, DBLP:journals/tbbis/CookHSTV19, DBLP:conf/fgr/HupontT19, DBLP:journals/corr/abs-1912-07398, DBLP:conf/btas/HowardSV19, DBLP:conf/cvpr/RobinsonLHQ0T20, FRVT2019, DBLP:journals/corr/abs-2007-06570} focused on jointly investigating the effects of user demographics on face recognition.
These studies showed that the effects lead to an exponential face recognition error increase when facing the same biased race, gender, and age factors \cite{DBLP:conf/btas/HowardSV19}.
Particular attention deserves the Face Recognition Vendor Test (FRVT) \cite{FRVT2019}, a large-scale benchmark of commercial algorithms analysing the face recognition performance with regards to demographics.
They consistently elevated false positives for female subjects and subjects at the outer ends of the age spectrum.
An overview of bias estimation in face recognition is shown in Table \ref{tab:SOTA}.

\begin{table}[]
\setlength{\tabcolsep}{4pt}
\centering
\caption{Overview of recent works analysing bias in face recognition. Identities and images refers to the used testing data. In contrast to previous works that analyses some specific demographic attributes, our work investigates a large range of demographic and non-demographic attributes.}
\label{tab:SOTA}
\begin{tabular}{lrrl}
\Xhline{2\arrayrulewidth} 
Work                      & Identities & Images & Attributes (number classes)                     \\
\hline
Ricanek et al. \cite{DBLP:conf/biosig/RicanekBS15}       & 0.7k       & 8.0k   & Age (2)                             \\
Deb et al. \cite{DBLP:conf/icb/DebN018}             & 0.9k       & 3.7k   & Age (cont.)                             \\
Srinivas et al. \cite{DBLP:conf/cvpr/SrinivasRMBK19}         & 1.7k       & 9.2k   & Age (2)                  \\
Michalski et al. \cite{DBLP:conf/icb/MichalskiYM18}       & -          & 4.7M   & Age (cont.)                             \\
Albiero et al. \cite{DBLP:journals/corr/abs-2008-06989}          & 26.9k      & 151.6k & Gender (2)                          \\
Albiero et al. \cite{DBLP:journals/corr/abs-2002-02934}          & 15.9k      & 101.3k & Gender (2)                          \\
Vera-Rodriguez et al. \cite{DBLP:conf/cvpr/Vera-RodriguezB19}  & 0.5k       & 169.4k & Gender (2)                          \\
Cavazos et al. \cite{DBLP:journals/corr/abs-1912-07398}          & 0.4k       & 1.1k   & Ethnicity (2)                       \\
Krishnapriya et al \cite{9001031}     & 22.7k      & 3.3M   & Gender (2), Ethnicity (2)               \\
Serna et al. \cite{DBLP:journals/corr/abs-2004-11246} & 55k & 1.4M & Gender (2), Ethnicity (4) \\
Acien et al. \cite{DBLP:conf/ciarp/AcienMVBF18} & 1.7k & 13k & Gender (2), Ethnicity (3)\\
Hupont et al. \cite{DBLP:conf/fgr/HupontT19}          & 0.6k       & 10.8k  & Gender, Ethnicity (3)               \\
Robinson et al. \cite{DBLP:conf/cvpr/RobinsonLHQ0T20}        & 0.8k       & 2.0k   & Gender (2), Ethnicity (4)               \\
Srinivas et al. \cite{8638319}        & 0.7k       & 8.0k   & Age (cont.), Gender (2)                     \\
Klare et al. \cite{6327355}           & 52.3k      & 102.9k & Age (3), Gender (2), \\
&&& Ethnicity (3)          \\
Howard et al. \cite{DBLP:conf/btas/HowardSV19}          & 1.1k       & 2.7k   & Age (cont.), Gender (2), \\
&&& Ethnicity (2)          \\
Grother et al. \cite{FRVT2019}         & 8.0M         & 18.0M    & Age (5), Gender (2), \\
&&& Ethnicity (4)          \\
Georgopoulos et al \cite{DBLP:journals/ivc/GeorgopoulosPP20} & 1.0k       & 41.0k  & Age (5), Gender (2), \\
&&& Kinship (5)            \\
Balakrishnan et al. \cite{DBLP:journals/corr/abs-2007-06570}     & 1.3k       & 1.3k   & Gender (2), Hair (cont.) \\
&&& Ethnicity (cont.)         \\
Cook et al. \cite{DBLP:journals/tbbis/CookHSTV19}            & 1.1k       & 2.7k   & Age (cont.), Gender  (2),  \\
&  &   & Ethnicity (4), Eyewear (2) \\
Lu et al. \cite{DBLP:journals/tbbis/LuCCC19} & 5.4k & 162.5k & Demographics (3),\\
 & & & Non-demographics (4)\\
\hline
This work                 & 9.1k       & 3.3M   & \textbf{Demographics (8)},\\
&&& \textbf{Non-demographics (40)}                  \\
\Xhline{2\arrayrulewidth}  
\end{tabular}
\end{table}

\vspace{-2mm}
\subsection{Mitigating Bias in Face Recognition}

The findings summarized in Section \ref{sec:RelatedWorkEstimatingBias} motivated research towards mitigating demographic-bias in face recognition approaches.
An early approach was presented by Zhang and Zhou \cite{Zhang:2010:CFR:1850487.1850593} who formulated the face verification problem as a multiclass cost-sensitive learning task and demonstrated that this approach can reduce different kinds of faulty decisions of the system.
In 2017, range loss \cite{DBLP:conf/iccv/ZhangFWLQ17} was proposed to learn robust face representations that can deal with long-tailed training data.
It is designed to reduce overall intrapersonal variations while enlarging interpersonal differences simultaneously.
Recent works aimed at mitigating demographic-bias in face recognition through adversarial learning \cite{gong2019debface,Liang_2019_CVPR,SensitiveNets, PrivacyNet}, margin-based approaches \cite{wang2019mitigate,DBLP:journals/corr/abs-1806-00194}, data augmentation \cite{WANG201935,Kortylewski_2019_CVPR_Workshops,DBLP:conf/cvpr/00010S0C19}, metric-learning \cite{DBLP:conf/iwbf/TerhorstD0K20}, or score normalization \cite{DBLP:journals/corr/abs-2002-03592}.

%

\vspace{-2mm}
\subsection{How This Work Contributes to the State of the Art}

So far, the majority of research in estimating and mitigating bias in face recognition focused on demographic factors such as age, gender, and race.
However, to achieve a generally accurate and fair face recognition model, it is necessary to know all potential origins of differential outcome.
Therefore, this work aims at closing this knowledge gap by analysing the differential outcome on a much wider attribute range than previous works (see Table \ref{tab:SOTA}).
More precisely, this work investigates the influence of 47 attributes on the face recognition performance of two popular face embeddings. 
The 47 attributes represent a step forward in the literature in comparison with previous analyses focused on no more than seven attributes \cite{DBLP:journals/tbbis/LuCCC19}.

\section{Experiments on Measuring Differential Outcome}

\subsection{Database and Considered Attributes}
To get reliable statements on the effect of different attributes on face recognition, we need a database that (a) provides a high number of face images with (b) many attribute annotations of (c) high quality.
For the experiments, we choose the publicly available MAAD-Face\footnote{\url{https://github.com/pterhoer/MAAD-Face}} annotation database \cite{maadface} based on the images of VGGFace2 \cite{Cao18} since this database fulfils our experimental requirements.
MAAD-Face provides over 120M high-quality attribute annotations of 3.3M face images of over 9k individuals.
It provides annotations for 47 distinct attributes of various kinds such as demographics, skin types, hair-styles and -colors, face geometry, annotations for the periocular, mouth, and nose area, as well as annotations for accessories.
An exact list of the MAAD-Face annotation attributes can be seen in Table \ref{tab:AttributePerformanceFaceNet1} and \ref{tab:AttributePerformanceFaceNet2}.
These attribute annotations proofed to have a higher quality than comparable face annotation databases \cite{maadface}.

\vspace{-2mm}
\subsection{Face Recognition Models}

For the experiments, we use two popular face recognition models, FaceNet \cite{DBLP:journals/corr/SchroffKP15} and ArcFace \cite{Deng_2019_CVPR}.
To create a face embedding for a given face image, the image has to be aligned, scaled, and cropped. 
Then, the preprocessed image is passed to a face recognition model to extract the embeddings.
For FaceNet, the preprocessing is done as described in \cite{Kazemi2014OneMF}.
To extract the embeddings, a pretrained model\footnote{\url{https://github.com/davidsandberg/facenet}} was used.
For ArcFace, the image preprocessing was done as described in \cite{DBLP:journals/corr/abs-1812-01936} and a pretrained model\footnote{\url{https://github.com/deepinsight/insightface}}  is used, which is provided by the authors of ArcFace.
Both models use a ResNet-100 architecture and were trained on the MS1M database \cite{DBLP:journals/corr/GuoZHHG16}.
The identity verification is done by comparing two embeddings with the widely-used cosine-similarity.

\vspace{-2mm}
\subsection{Evaluation Metrics}
\label{sec:EvaluationMetrics}

The face verification performance is reported in terms of (a) false non-match rates (FNMR) at a fixed false match rate (FMR) and (b) equal error rates (EER).
The EER equals the FMR at the threshold where FMR = FNMR and is well known as a single-value indicator of the verification performance.
The used error rates are specified for biometric verification evaluation in the international standard \cite{ISO_Metrik}.
In the experiments, the face verification performance is reported on three operating points to cover a wide range of potential applications.
This includes EER, as well as, the FNMR at $10^{-3}$ and $10^{-4}$ FMR as recommended by the best practice guidelines for automated border control of the European Boarder Guard Agency Frontex \cite{FrontexBestPractice}.
For each operating point and attribute, the verification performance is computed on all samples with positive and all samples with negative annotations.
This will allow to compare the performance differences of face embeddings regarding binary attributes, such as bald vs non-bald faces.

\vspace{-2mm}
\subsection{Control Groups}
\label{sec:ControlGroups}

During the experiments, the number of testing samples with positive and negative labels might be significantly different.
To prevent misleading conclusions from such unbalanced annotation distributions, we introduce positive and negative control groups for each attribute.
For each attribute, six positive and six negative control groups are created by randomly selecting samples from the database.
This control group creation is done such that the synthetic control groups have the same number of samples as their positive and negative counterparts.

Comparing the verification performance of the positive and negative control groups allow us to state the validity of the (real) attribute-based verification performance.
If the performances of the negative and positive control groups is very similar, the (real) attribute recognition performance is treated as valid.
In this case, the unbalanced testing data distribution shows no effect on the performance.
If the relative performance of the control groups differ strongly, the recognition performance might be significantly affected from the unbalanced distribution of the positively and negatively annotated samples.
In this case, the (real) attribute recognition performance might be affected as well.
Consequently, statements about the influence of this attribute on the recognition performance are of low validity.
In the experiments, the validity $val$ of an attribute $a$
\begin{align}
val(a) = 1 - \dfrac{err_{control}^{(+)}(a)}{err_{control}^{(-)}(a)}, \label{eq:validity}
\end{align}
is defined over the relative performance differences between the control groups.
The terms $err_{control}^{(+)}(a)$ and $err_{control}^{(-)}(a)$ represent the recognition errors of the positive $(+)$ and the negative $(-)$ control groups of attribute $a$.
For the experiments, we consider attributes with a validity of $<0.9$ as \textit{not valid}.
However, we will also present the performance differences with the corresponding validity values so that the operators are able to choose more suitable validity threshold for their applications.

\vspace{-2mm}
\subsection{Investigations}

To analyse the influence of different attributes on the recognition performance of two face recognition models, the investigations are divided into several parts.

\begin{enumerate}
\item To emphasize if an attribute bias might originate from correlated attribute annotations, we analyse the correlation between the attribute annotations.
\item For each attribute, the recognition performance of its positively- and negatively-labelled attribute groups are compared to investigate the influence of this attribute on the recognition performance.
The results are discussed in the context of the corresponding validity values to avoid misinterpretations occurring from unbalanced testing data.
\item A visual summary is provided to relate the impact of the attributes on the face recognition systems to the validity of the results.
This aims at providing a compact and easily-understandable overview of the findings of this work.
\item We provide possible explanations causing the differential outcome and discuss these differences between both face recognition systems. 
\item Lastly, we use the observations to derive future research directions for face recognition systems. 
\end{enumerate}

\vspace{-2mm}

\section{Results}

\subsection{Investigating the Correlation of Facial Attributes}
\label{sec:AttributeCorrelation}

\begin{figure}
\centering
\includegraphics[width=0.4\textwidth]{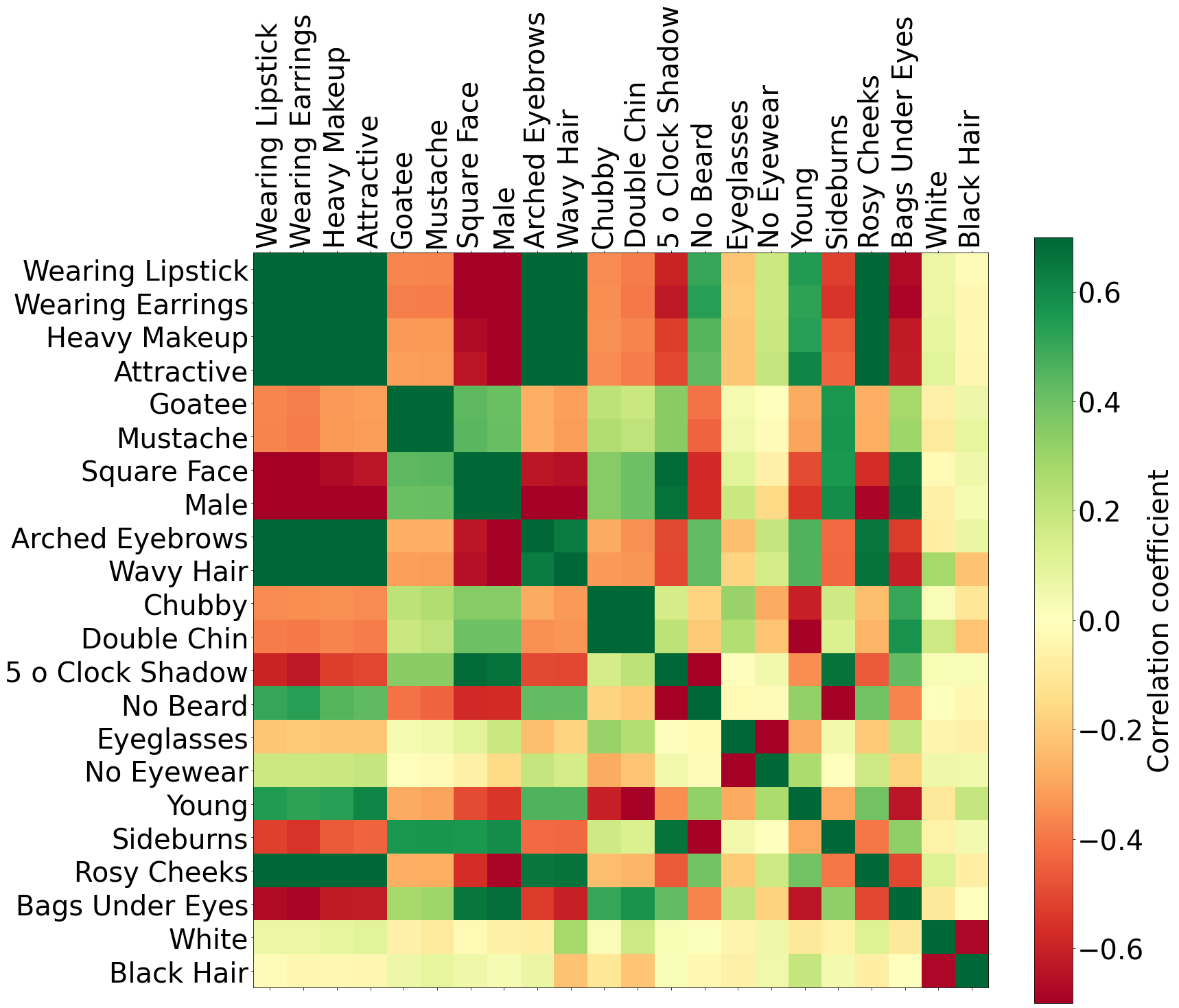}
\caption{Compressed annotation correlations of the used MAAD-Face database. The attributes are chosen such that the 15 most positive and negative pairwise are visible. Green indicate positive
correlations, while red indicate a negative correlation. The correlation is based on the Pearson coefficient. When interpreting the results from Section \ref{sec:Results_Detailed} highly-correlated attributes should be considered to prevent misinterpretations. 
\vspace{-5mm}
}
\label{fig:CorrelationMatrixSmall}
\end{figure}

To understand the quality of the used labels and potential biases in the attribute space, Figure \ref{fig:CorrelationMatrixSmall} shows a selection of specific attribute-label correlations.
The attributes are chosen to show the 15 most positive and negative pairwise correlations.
It can be seen that \textit{Wearing Lipstick}, \textit{Wearing Earrings}, \textit{Heavy Makeup}, \textit{Young}, and \textit{Attractive} correlates highly positively with \textit{Arched Eyebrows}, \textit{Wavy Hair}, and \textit{Rosy Cheeks}.
In contrast, these attributes correlates negatively with \textit{Square Face}, \textit{Male}, and \textit{Bags Under Eyes}.
These correlations have to be considered when comparing the differential outcome for the different attributes.
However, the correlation matrix also approves the quality of some labels that semantically excludes each other.
For instance, \textit{5 o Clock Shadow} negatively correlates with \textit{No Beard} and \textit{Eyeglasses} negatively correlates with \textit{No Eyewear}.

\subsection{The Impact of Facial Attributes on Recognition}
\label{sec:Results_Detailed}

The main contribution of this work is an analysis of the effect of 47 distinct attributes on two popular face recognition models.
This aims at investigating model biases.
For each attribute, the face verification performance is calculated on positively-labelled samples, as well as on negatively-labelled samples. 
This is done on three operating points as explained in Section \ref{sec:EvaluationMetrics}.
The relative performance between the positive and negative groups allows to investigate potential biases of the face recognition model towards the analysed attribute.
To determine if differential outcome results from unbalanced data distributions we introduced control groups as explained in Section \ref{sec:ControlGroups}.

In Tables \ref{tab:AttributePerformanceFaceNet1}, \ref{tab:AttributePerformanceFaceNet2}, \ref{tab:AttributePerformanceArcFace1}, and \ref{tab:AttributePerformanceArcFace2} the performance of the positive and negative class is shown for each attribute.
The performance of the annotated data is referred as Real while the performance of the control groups is referred as Control.
The relative performance (Rel. Perf.) shows the relative performance difference between the positive and negative attribute classes.
If the relative performances between the control classes are below 10\% ($val \geq 0.9)$, the result is considered as \textit{valid} (green highlighting).
Otherwise, the result is considered as \textit{not valid} indicated by a grey highlighting.
Positive values for the relative performance of an attribute represents a positive effect of the attribute on the face recognition performance.
Negative values indicate a negative influence of the attribute on the recognition performance. 
In the following, we present the results of our study on bias on FaceNet and ArcFace embeddings.

\subsubsection{Biases in FaceNet Embeddings}

The results of our attribute-related study on differential outcome of the FaceNet model are shown in Table \ref{tab:AttributePerformanceFaceNet1} and \ref{tab:AttributePerformanceFaceNet2}.

Previous works focused on differential outcome affected by the user's demographics.
The results on FaceNet confirms the observations of these works.
Demographics strongly affect the recognition performance.
One of the strongest impacts on FaceNet is observed for ethnicities. 
For the investigated FaceNet model, \textit{Asian} and \textit{Black} faces lead to significantly lower recognition rates than \textit{White} faces.
Also \textit{Young} ones perform significantly weaker than \textit{Middle-aged} faces.
Concerning gender, we observe that \textit{Male} face perform better then \textit{Female} ones.
These finding are intensively discussed in previous works \cite{DBLP:journals/corr/abs-2008-06989, DBLP:journals/corr/abs-2002-02934}.
However, the experimental results show that there are many more aspects that strongly affect the recognition performance.

One factor leading to differential outcome is the user's hair.
While \textit{Bald} faces and \textit{Receding Hairlines} lead to a improved recognition performance, \textit{Wavy Hair} styles or \textit{Bangs} are observed to degrade the performance.
This can be explained by the visibility of the face.
In general, \textit{Wavy Hair} and \textit{Bangs} are more likely to cover parts of the face while \textit{Bald} faces or faces with \textit{Receding Hairlines} do not occlude part of the faces.

A contradictory observation can be made for facial hair.
Faces with \textit{No-Beard} perform worse than faces with a beard, such as a \textit{5 o Clock Shadow}.
A reason for this can be that people might keep their beards over a long period of time and thus, the training and testing data might be biased such that the recognition networks consider the beards for recognition.
 
Also the color of the hair has an impact on the FaceNet embeddings.
While \textit{Blond Hair} shows a strongly degraded face recognition performance, \textit{Gray Hair} leads to the strongest performances.

The results indicate that the shape of a face only have a minor impact on the face recognition performance.
For \textit{Oval Faces}, no significant differences to non-oval faces could be observed.
Although, a positive effect on the recognition performance is shown for \textit{Square Faces}, in Section  \ref{sec:AttributeCorrelation} a strong correlation between \textit{Square Face} and \textit{Male} was shown.
This might explain the behaviour.

Faces with \textit{High Cheekbones}, \textit{Double Chins}, and \textit{Chubby} faces also perform better for FaceNet features than the inverted counterparts.
Probably because these properties provide additional information that can be used for recognition.
In contrast to this, an \textit{Obstructed Forehead} strongly degrades the recognition performance while a \textit{Fully Visible Forehead} provides additional (uncovered) information that supports the recognition process.

Anomalous properties in the periocular area, such as \textit{Bags Under Eyes}, \textit{Bushy Eyebrows}, or \textit{Arched Eyebrows}, lead to better recognition rates compared to face images without these attributes. 
The same goes for \textit{Big Nose} and \textit{Pointy Nose}.

The reason that \textit{Smiling} and a \textit{Mouth Closed} lead to a stronger recognition performances than non-neutral expressions might be explainable through the used face databases that mainly contain of face with these expression.
In \cite{DBLP:conf/btas/DamerWBBT0K18}, the opposite effect was already shown by demonstrating that crazy faces result in low comparison scores.
However, this considered extreme expressions aim at avoiding identification.

Interestingly, accessories have a strong influence on the recognition performance of FaceNet.
\textit{Wearing Hat}, \textit{Wearing Earrings}, or \emph{Eyeglasses} degrade the face recognition performance significantly and might be explained by the fact that these accessories cover parts of the face.

\subsubsection{Biases in ArcFace Embeddings}

The results of our attribute-related study on differential outcome of the ArcFace model are shown in Table \ref{tab:AttributePerformanceArcFace1} and \ref{tab:AttributePerformanceArcFace2}.

Similar to FaceNet, the results on ArcFace confirms the observed demographic performance differences shown by previous works.
For the investigated ArcFace model, \textit{Young} faces perform weaker than  \textit{Middle-aged} or \textit{Senior} faces.
Interestingly, the intensively discussed gender bias is strongly dependent on the used decision threshold.
Especially for lower FMRs the differential outcome between \textit{Male} and \textit{Female} increases.
Concerning the ethnic-bias on the ArcFace model, we are not able to confirm the observations from previous works.
For \textit{White} faces, the performance is significantly higher than for non-white faces.
For \textit{Asian} and \textit{Black} faces, a strong degradation in the recognition performance can be observed.
However, we have to consider this results as \textit{not valid}, since we can observe strong performance differences on the control groups.
This indicates that these results are strongly influenced by the unbalanced testing data.

Similar to FaceNet, the user's hair shows to have a significant impact on the face recognition performance.
While, \textit{Receding Hairlines}, \textit{Wavy Hair} and \textit{Sideburns} supports the recognition process, faces with \textit{Bangs} show a strong degradation.
Again, the performance differences on ArcFace show to be threshold-dependent.
For \textit{Wavy Hair}, the positive effect on face recognition vanishes for lower FMRs, and for \textit{Bangs}, the negative effect increases drastically for lower FMRs.

Also the color of the user's hair have an impact on the recognition performance.
\textit{Gray Hair} performs significantly above average, while \textit{Black Hair} performs significantly below average.
\textit{Blond Hair} and \textit{Brown Hair} lead to differential outcome depending on the decision threshold.
For high FMRs, \textit{Blond Hair} improves the recognition performance, while for lower FMRs, the recognition performance changes to below-average.
For faces with \textit{Brown Hair}, the positive effect on recognition vanishes for lower FMRs.

The effect of wearing a beard on the performance of ArcFace is similar to FaceNet.
Having \textit{No Beard} decreases the recognition performance and having a beard, such as a \textit{5 o Clock Shadow}, enhances the recognition.
These effects are clearer for lower FMRs.

On contrast to FaceNet, the face shape effects the recognition performance of ArcFace.
Both, \textit{Oval Faces} and \textit{Square Faces} have positive effect on the recognition performance, which is dependent on the utilized decision threshold.
\textit{Round Faces} show a strongly degraded recognition.
However, a large fraction of this performance differences can be explained by the unbalanced data distribution and thus, we have to neglect the results for \textit{Round Faces}.

Similar to FaceNet, \textit{High Cheekbones}, \textit{Double Chin}, \textit{Chubby}, and a \textit{Fully Visible Forehead} lead to improved face recognition performances.
While a \textit{Fully Visible Forehead} refers to no partial occlusions of the face that might negatively infer, the other attributes provide anomalous characteristics that might help for recognition. 

Surprisingly, faces with \textit{Brown Eyes} perform drastically weaker than faces with non-brown eyes.
For \textit{Bags Under Eyes}, \textit{Bushy Eyebrows}, and \textit{Arched Eyebrows}, an improved face recognition performance can be observed.
These attributes can be treated as anomalies and thus, can support the recognition process.
The same goes for \textit{Big Nose} and \textit{Pointy Nose}.


Similar to FaceNet, accessories have a strong impact on the differential outcome of ArcFace.
While having \textit{Heavy Makeup}, such as \textit{Wearing Lipstick}, improves the recognition, faces with \textit{Eyeglasses} or \textit{Wearing Hat} lead to strong degradations in the face recognition performance. 
A reason for this might be that people using \textit{Heavy Makeup} frequently.
Consequently, a person in the training data might either have no or only \textit{Heavy Makeup} images.
On the other side, people tend to change their \textit{Eyeglasses} or \textit{(Wearing) Hats} more frequently.
Moreover, these attributes might lead to partial occlusions of the face leading to less identity-information available and thus, to a degraded face recognition performance.




\begin{table*}[]
\renewcommand{\arraystretch}{1.0}
\centering
\caption{FaceNet - Part 1/2. Face recognition performance based on several attributes.}
\label{tab:AttributePerformanceFaceNet1}
\begin{tabular}{lllrrrrrr}
\Xhline{2\arrayrulewidth}
Category              & Attribute                         & Class  & \multicolumn{2}{c}{EER} & \multicolumn{2}{c}{FNMR@FMR=$10^{-3}$} & \multicolumn{2}{c}{FNMR@FMR=$10^{-4}$} \\
\cmidrule(rl){4-5} \cmidrule(rl){6-7} \cmidrule(rl){8-9}
              &                          &            & \multicolumn{1}{c}{Real}        & \multicolumn{1}{c}{Control}   & \multicolumn{1}{c}{Real}               & \multicolumn{1}{c}{Control}          & \multicolumn{1}{c}{Real}               & \multicolumn{1}{c}{Control}          \\
              \hline
Demographics  &  Male                     & Positive   & 6.64\%      & 6.49\%    & 33.28\%            & 32.51\%          & 53.64\%            & 52.44\%          \\
             &                          & Negative   & 7.87\%      & 6.46\%    & 42.47\%            & 32.40\%          & 62.55\%            & 52.32\%          \\
\rowcolor{PaleGreen2} \cellcolor{White} & \cellcolor{White}     & Rel. Perf. & 15.56\%     & -0.42\%   & 21.63\%            & -0.35\%          & 14.24\%            & -0.21\%          \\
              & Young                    & Positive   & 6.91\%      & 6.46\%    & 39.39\%            & 32.39\%          & 60.37\%            & 52.30\%          \\
              &                          & Negative   & 5.73\%      & 6.47\%    & 28.93\%            & 32.37\%          & 48.97\%            & 52.18\%          \\
    \rowcolor{PaleGreen2} \cellcolor{White} & \cellcolor{White}                         & Rel. Perf. & -20.58\%    & 0.12\%    & -36.19\%           & -0.08\%          & -23.27\%           & -0.22\%          \\
              & Middle\_Aged             & Positive   & 5.41\%      & 6.33\%    & 28.77\%            & 31.70\%          & 48.75\%            & 51.25\%          \\
              &                          & Negative   & 6.96\%      & 6.48\%    & 36.77\%            & 32.52\%          & 57.70\%            & 52.45\%          \\
\rowcolor{PaleGreen2} \cellcolor{White} & \cellcolor{White} & Rel. Perf. & 22.29\%     & 2.33\%    & 21.74\%            & 2.52\%           & 15.52\%            & 2.28\%           \\
              & Senior                   & Positive   & 6.01\%      & 6.23\%    & 30.26\%            & 31.19\%          & 50.52\%            & 50.52\%          \\
              &                          & Negative   & 6.69\%      & 6.49\%    & 34.19\%            & 32.54\%          & 54.58\%            & 52.53\%          \\
\rowcolor{PaleGreen2} \cellcolor{White} & \cellcolor{White} & Rel. Perf. & 10.16\%     & 3.93\%    & 11.50\%            & 4.16\%           & 7.44\%             & 3.82\%           \\
              & Asian                    & Positive   & 11.16\%     & 5.91\%    & 69.46\%            & 29.52\%          & 88.48\%            & 48.20\%          \\
              &                          & Negative   & 6.33\%      & 6.49\%    & 31.91\%            & 32.55\%          & 51.27\%            & 52.54\%          \\
\rowcolor{PaleGreen2} \cellcolor{White} & \cellcolor{White} & Rel. Perf. & -76.30\%    & 8.88\%    & -117.66\%          & 9.33\%           & -72.58\%           & 8.28\%           \\
              & White                    & Positive   & 5.97\%      & 6.48\%    & 31.28\%            & 32.51\%          & 50.15\%            & 52.50\%          \\
              &                          & Negative   & 7.51\%      & 6.44\%    & 46.82\%            & 32.16\%          & 69.44\%            & 51.94\%          \\
\rowcolor{PaleGreen2} \cellcolor{White} & \cellcolor{White} & Rel. Perf. & 20.54\%     & -0.56\%   & 33.18\%            & -1.11\%          & 27.79\%            & -1.07\%          \\
              & Black                    & Positive   & 8.85\%      & 6.02\%    & 52.50\%            & 30.20\%          & 73.61\%            & 49.32\%          \\
              &                          & Negative   & 6.61\%      & 6.49\%    & 33.47\%            & 32.54\%          & 53.34\%            & 52.52\%          \\
\rowcolor{PaleGreen2} \cellcolor{White} & \cellcolor{White} & Rel. Perf. & -33.98\%    & 7.14\%    & -56.89\%           & 7.19\%           & -37.99\%           & 6.09\%           \\
Skin          & Rosy\_Cheeks             & Positive   & 1.29\%      & 5.46\%    & 3.76\%             & 26.05\%          & 9.46\%             & 42.72\%          \\
              &                          & Negative   & 7.36\%      & 6.48\%    & 37.03\%            & 32.51\%          & 57.65\%            & 52.47\%          \\
\rowcolor{Snow2} \cellcolor{White} & \cellcolor{White} & Rel. Perf. & 82.42\%     & 15.80\%   & 89.86\%            & 19.87\%          & 83.59\%            & 18.59\%          \\
              & Shiny\_Skin              & Positive   & 6.08\%      & 6.41\%    & 36.43\%            & 32.05\%          & 57.46\%            & 51.83\%          \\
              &                          & Negative   & 7.90\%      & 6.47\%    & 41.33\%            & 32.37\%          & 62.43\%            & 52.29\%          \\
\rowcolor{PaleGreen2} \cellcolor{White} & \cellcolor{White} & Rel. Perf. & 23.06\%     & 0.85\%    & 11.86\%            & 0.99\%           & 7.97\%             & 0.88\%           \\
Hair          & Bald                     & Positive   & 5.10\%      & 6.13\%    & 30.52\%            & 30.69\%          & 52.37\%            & 49.93\%          \\
              &                          & Negative   & 6.70\%      & 6.49\%    & 34.13\%            & 32.54\%          & 54.47\%            & 52.52\%          \\
\rowcolor{PaleGreen2} \cellcolor{White} & \cellcolor{White} & Rel. Perf. & 23.89\%     & 5.55\%    & 10.59\%            & 5.70\%           & 3.85\%             & 4.94\%           \\
              & Wavy\_Hair               & Positive   & 7.55\%      & 6.46\%    & 40.97\%            & 32.29\%          & 60.69\%            & 52.10\%          \\
              &                          & Negative   & 6.82\%      & 6.48\%    & 34.23\%            & 32.50\%          & 54.84\%            & 52.48\%          \\
\rowcolor{PaleGreen2} \cellcolor{White} & \cellcolor{White} & Rel. Perf. & -10.68\%    & 0.34\%    & -19.69\%           & 0.65\%           & -10.65\%           & 0.73\%           \\
              & Receding\_Hairline       & Positive   & 4.93\%      & 6.43\%    & 26.02\%            & 32.06\%          & 44.95\%            & 51.75\%          \\
              &                          & Negative   & 7.35\%      & 6.47\%    & 39.92\%            & 32.46\%          & 61.12\%            & 52.43\%          \\
\rowcolor{PaleGreen2} \cellcolor{White} & \cellcolor{White} & Rel. Perf. & 32.90\%     & 0.74\%    & 34.82\%            & 1.23\%           & 26.46\%            & 1.29\%           \\
              & Bangs                    & Positive   & 6.82\%      & 6.34\%    & 45.53\%            & 31.63\%          & 69.28\%            & 51.14\%          \\
              &                          & Negative   & 6.43\%      & 6.49\%    & 32.02\%            & 32.54\%          & 51.86\%            & 52.52\%          \\
\rowcolor{PaleGreen2} \cellcolor{White} & \cellcolor{White} & Rel. Perf. & -5.99\%     & 2.25\%    & -42.17\%           & 2.79\%           & -33.59\%           & 2.62\%           \\
              & Sideburns                & Positive   & 6.68\%      & 6.46\%    & 34.33\%            & 32.34\%          & 54.08\%            & 52.23\%          \\
              &                          & Negative   & 6.75\%      & 6.49\%    & 35.47\%            & 32.52\%          & 55.96\%            & 52.46\%          \\
\rowcolor{PaleGreen2} \cellcolor{White} & \cellcolor{White} & Rel. Perf. & 1.04\%      & 0.43\%    & 3.24\%             & 0.53\%           & 3.35\%             & 0.44\%           \\
              & Black\_Hair              & Positive   & 7.13\%      & 6.42\%    & 42.35\%            & 32.06\%          & 65.73\%            & 51.73\%          \\
              &                          & Negative   & 6.20\%      & 6.48\%    & 32.46\%            & 32.49\%          & 52.06\%            & 52.47\%          \\
\rowcolor{PaleGreen2} \cellcolor{White} & \cellcolor{White} & Rel. Perf. & -15.04\%    & 1.02\%    & -30.47\%           & 1.34\%           & -26.26\%           & 1.40\%           \\
              & Blond\_Hair              & Positive   & 9.63\%      & 6.34\%    & 52.00\%            & 31.66\%          & 71.71\%            & 51.18\%          \\
              &                          & Negative   & 6.45\%      & 6.48\%    & 32.63\%            & 32.52\%          & 52.96\%            & 52.48\%          \\
\rowcolor{PaleGreen2} \cellcolor{White} & \cellcolor{White} & Rel. Perf. & -49.35\%    & 2.17\%    & -59.37\%           & 2.66\%           & -35.41\%           & 2.48\%           \\
              & Brown\_Hair              & Positive   & 7.40\%      & 6.45\%    & 39.73\%            & 32.26\%          & 59.12\%            & 52.06\%          \\
              &                          & Negative   & 6.19\%      & 6.47\%    & 35.13\%            & 32.41\%          & 57.09\%            & 52.30\%          \\
\rowcolor{PaleGreen2} \cellcolor{White} & \cellcolor{White} & Rel. Perf. & -19.52\%    & 0.26\%    & -13.08\%           & 0.49\%           & -3.55\%            & 0.48\%           \\
              & Gray\_Hair               & Positive   & 5.32\%      & 6.29\%    & 26.00\%            & 31.50\%          & 44.11\%            & 50.99\%          \\
              &                          & Negative   & 6.72\%      & 6.49\%    & 34.60\%            & 32.54\%          & 55.25\%            & 52.52\%          \\
\rowcolor{PaleGreen2} \cellcolor{White} & \cellcolor{White} & Rel. Perf. & 20.83\%     & 3.05\%    & 24.83\%            & 3.20\%           & 20.17\%            & 2.90\%           \\
Beard         & No\_Beard                & Positive   & 7.20\%      & 6.48\%    & 37.97\%            & 32.49\%          & 58.83\%            & 52.44\%          \\
              &                          & Negative   & 6.13\%      & 6.40\%    & 31.07\%            & 31.94\%          & 51.01\%            & 51.60\%          \\
\rowcolor{PaleGreen2} \cellcolor{White} & \cellcolor{White} & Rel. Perf. & -17.53\%    & -1.38\%   & -22.20\%           & -1.74\%          & -15.33\%           & -1.62\%          \\
              & Mustache                 & Positive   & 6.45\%      & 4.93\%    & 50.77\%            & 22.55\%          & 73.71\%            & 36.74\%          \\
              &                          & Negative   & 6.90\%      & 6.48\%    & 35.54\%            & 32.52\%          & 56.12\%            & 52.49\%          \\
\rowcolor{Snow2} \cellcolor{White} & \cellcolor{White} & Rel. Perf. & 6.41\%      & 23.94\%   & -42.88\%           & 30.68\%          & -31.34\%           & 30.01\%          \\
              & 5\_o\_Clock\_Shadow      & Positive   & 6.16\%      & 6.38\%    & 30.98\%            & 31.87\%          & 50.01\%            & 51.53\%          \\
              &                          & Negative   & 7.49\%      & 6.48\%    & 39.91\%            & 32.46\%          & 60.86\%            & 52.39\%          \\
\rowcolor{PaleGreen2} \cellcolor{White} & \cellcolor{White} & Rel. Perf. & 17.78\%     & 1.54\%    & 22.37\%            & 1.83\%           & 17.83\%            & 1.64\%           \\
              & Goatee                   & Positive   & 2.59\%      & 4.69\%    & 18.78\%            & 20.11\%          & 38.17\%            & 32.98\%          \\
              &                          & Negative   & 6.92\%      & 6.49\%    & 35.49\%            & 32.54\%          & 56.11\%            & 52.55\%          \\
\rowcolor{Snow2} \cellcolor{White} & \cellcolor{White} & Rel. Perf. & 62.59\%     & 27.63\% & 47.09\%            & 38.19\%          & 31.97\%            & 37.24\%          \\
\Xhline{2\arrayrulewidth}     
\end{tabular}
\end{table*}

\begin{table*}[]
\renewcommand{\arraystretch}{1.0}
\centering
\caption{FaceNet - Part 2/2. Face recognition performance based on several attributes.}
\label{tab:AttributePerformanceFaceNet2}
\begin{tabular}{lllrrrrrr}
\Xhline{2\arrayrulewidth}
Category              & Attribute                         & Class  & \multicolumn{2}{c}{EER} & \multicolumn{2}{c}{FNMR@FMR=$10^{-3}$} & \multicolumn{2}{c}{FNMR@FMR=$10^{-4}$} \\
\cmidrule(rl){4-5} \cmidrule(rl){6-7} \cmidrule(rl){8-9}
              &                          &            & \multicolumn{1}{c}{Real}        & \multicolumn{1}{c}{Control}   & \multicolumn{1}{c}{Real}               & \multicolumn{1}{c}{Control}          & \multicolumn{1}{c}{Real}               & \multicolumn{1}{c}{Control}          \\
              \hline
Face Geometry & Oval\_Face               & Positive   & 8.14\%      & 6.40\%    & 45.16\%            & 31.97\%          & 64.96\%            & 51.64\%          \\
              &                          & Negative   & 8.26\%      & 6.46\%    & 45.11\%            & 32.30\%          & 67.44\%            & 52.08\%          \\
\rowcolor{PaleGreen2} \cellcolor{White} & \cellcolor{White} & Rel. Perf. & 1.45\%      & 1.01\%    & -0.11\%            & 1.00\%           & 3.68\%             & 0.84\%           \\
              & Square\_Face             & Positive   & 6.32\%      & 6.48\%    & 31.37\%            & 32.49\%          & 51.25\%            & 52.44\%          \\
              &                          & Negative   & 7.81\%      & 6.47\%    & 41.51\%            & 32.43\%          & 61.90\%            & 52.38\%          \\
\rowcolor{PaleGreen2} \cellcolor{White} & \cellcolor{White} & Rel. Perf. & 19.13\%     & -0.12\%   & 24.42\%            & -0.16\%          & 17.20\%            & -0.12\%          \\
              & Round\_Face              & Positive   & 16.53\%     & 4.52\%    & 88.11\%            & 19.03\%          & 93.33\%            & 31.40\%          \\
              &                          & Negative   & 5.31\%      & 6.49\%    & 27.06\%            & 32.52\%          & 45.05\%            & 52.46\%          \\
\rowcolor{Snow2} \cellcolor{White} & \cellcolor{White} & Rel. Perf. & -211.14\%   & 30.27\%   & -225.65\%          & 41.49\%          & -107.17\%          & 40.14\%          \\
              & Double\_Chin             & Positive   & 5.45\%      & 6.44\%    & 26.28\%            & 32.15\%          & 44.43\%            & 51.85\%          \\
              &                          & Negative   & 7.09\%      & 6.48\%    & 38.20\%            & 32.51\%          & 59.43\%            & 52.46\%          \\
\rowcolor{PaleGreen2} \cellcolor{White} & \cellcolor{White} & Rel. Perf. & 23.05\%     & 0.71\%    & 31.20\%            & 1.10\%           & 25.24\%            & 1.18\%           \\
              & High\_Cheekbones         & Positive   & 5.99\%      & 6.46\%    & 33.69\%            & 32.27\%          & 53.73\%            & 52.10\%          \\
              &                          & Negative   & 8.10\%      & 6.47\%    & 41.66\%            & 32.41\%          & 62.29\%            & 52.32\%          \\
\rowcolor{PaleGreen2} \cellcolor{White} & \cellcolor{White} & Rel. Perf. & 26.11\%     & 0.20\%    & 19.13\%            & 0.43\%           & 13.73\%            & 0.43\%           \\
              & Chubby                   & Positive   & 5.11\%      & 6.38\%    & 26.98\%            & 31.81\%          & 47.76\%            & 51.48\%          \\
              &                          & Negative   & 6.85\%      & 6.49\%    & 36.65\%            & 32.54\%          & 57.76\%            & 52.49\%          \\
\rowcolor{PaleGreen2} \cellcolor{White} & \cellcolor{White} & Rel. Perf. & 25.35\%     & 1.62\%    & 26.38\%            & 2.23\%           & 17.31\%            & 1.92\%           \\
              & Obstructed\_Forehead     & Positive   & 8.85\%      & 6.11\%    & 60.01\%            & 30.67\%          & 80.51\%            & 49.92\%          \\
              &                          & Negative   & 6.02\%      & 6.49\%    & 31.14\%            & 32.52\%          & 50.70\%            & 52.50\%          \\
\rowcolor{PaleGreen2} \cellcolor{White} & \cellcolor{White} & Rel. Perf. & -46.87\%    & 5.75\%    & -92.69\%           & 5.69\%           & -58.79\%           & 4.91\%           \\
              & Fully\_Visible\_Forehead & Positive   & 5.47\%      & 6.48\%    & 28.25\%            & 32.46\%          & 47.36\%            & 52.35\%          \\
              &                          & Negative   & 7.82\%      & 6.45\%    & 44.34\%            & 32.28\%          & 66.70\%            & 52.09\%          \\
\rowcolor{PaleGreen2} \cellcolor{White} & \cellcolor{White} & Rel. Perf. & 30.01\%     & -0.43\%   & 36.29\%            & -0.55\%          & 28.99\%            & -0.49\%          \\
Periocular    & Brown\_Eyes              & Positive   & 7.54\%      & 6.48\%    & 42.04\%            & 32.44\%          & 63.89\%            & 52.36\%          \\
              &                          & Negative   & 6.12\%      & 6.36\%    & 33.59\%            & 31.83\%          & 52.03\%            & 51.50\%          \\
\rowcolor{PaleGreen2} \cellcolor{White} & \cellcolor{White} & Rel. Perf. & -23.28\%    & -1.81\%   & -25.15\%           & -1.94\%          & -22.78\%           & -1.67\%          \\
              & Bags\_Under\_Eyes        & Positive   & 5.90\%      & 6.45\%    & 31.51\%            & 32.31\%          & 52.50\%            & 52.16\%          \\
              &                          & Negative   & 8.03\%      & 6.47\%    & 42.47\%            & 32.42\%          & 62.85\%            & 52.31\%          \\
\rowcolor{PaleGreen2} \cellcolor{White} & \cellcolor{White} & Rel. Perf. & 26.47\%     & 0.36\%    & 25.79\%            & 0.37\%           & 16.48\%            & 0.29\%           \\
              & Bushy\_Eyebrows          & Positive   & 5.66\%      & 6.47\%    & 29.86\%            & 32.36\%          & 49.67\%            & 52.29\%          \\
              &                          & Negative   & 7.26\%      & 6.48\%    & 37.79\%            & 32.51\%          & 58.28\%            & 52.45\%          \\
\rowcolor{PaleGreen2} \cellcolor{White} & \cellcolor{White} & Rel. Perf. & 22.03\%     & 0.23\%    & 21.00\%            & 0.44\%           & 14.77\%            & 0.31\%           \\
              & Arched\_Eyebrows         & Positive   & 5.99\%      & 6.46\%    & 33.71\%            & 32.28\%          & 52.99\%            & 52.06\%          \\
              &                          & Negative   & 7.59\%      & 6.48\%    & 38.64\%            & 32.47\%          & 59.96\%            & 52.40\%          \\
\rowcolor{PaleGreen2} \cellcolor{White} & \cellcolor{White} & Rel. Perf. & 21.10\%     & 0.37\%    & 12.75\%            & 0.58\%           & 11.62\%            & 0.64\%           \\
Mouth         & Mouth\_Closed            & Positive   & 5.25\%      & 5.99\%    & 27.84\%            & 29.97\%          & 46.77\%            & 48.87\%          \\
              &                          & Negative   & 7.05\%      & 6.41\%    & 46.08\%            & 32.00\%          & 68.38\%            & 51.71\%          \\
\rowcolor{PaleGreen2} \cellcolor{White} & \cellcolor{White} & Rel. Perf. & 25.49\%     & 6.53\%    & 39.60\%            & 6.34\%           & 31.60\%            & 5.50\%           \\
              & Smiling                  & Positive   & 6.08\%      & 6.44\%    & 34.06\%            & 32.17\%          & 53.51\%            & 51.91\%          \\
              &                          & Negative   & 8.67\%      & 6.46\%    & 47.88\%            & 32.36\%          & 70.12\%            & 52.23\%          \\
\rowcolor{PaleGreen2} \cellcolor{White} & \cellcolor{White} & Rel. Perf. & 29.86\%     & 0.28\%    & 28.87\%            & 0.58\%           & 23.68\%            & 0.61\%           \\
              & Big\_Lips                & Positive   & 6.79\%      & 6.45\%    & 39.95\%            & 32.33\%          & 61.39\%            & 52.19\%          \\
              &                          & Negative   & 6.97\%      & 6.47\%    & 34.09\%            & 32.44\%          & 53.99\%            & 52.36\%          \\
\rowcolor{PaleGreen2} \cellcolor{White} & \cellcolor{White} & Rel. Perf. & 2.58\%      & 0.32\%    & -17.20\%           & 0.31\%           & -13.72\%           & 0.31\%           \\
Nose          & Big\_Nose                & Positive   & 6.28\%      & 6.42\%    & 36.68\%            & 32.04\%          & 59.22\%            & 51.82\%          \\
              &                          & Negative   & 8.40\%      & 6.48\%    & 46.15\%            & 32.43\%          & 67.05\%            & 52.32\%          \\
\rowcolor{PaleGreen2} \cellcolor{White} & \cellcolor{White} & Rel. Perf. & 25.23\%     & 0.90\%    & 20.52\%            & 1.20\%           & 11.67\%            & 0.94\%           \\
              & Pointy\_Nose             & Positive   & 6.04\%      & 6.48\%    & 32.67\%            & 32.48\%          & 51.66\%            & 52.44\%          \\
              &                          & Negative   & 7.80\%      & 6.46\%    & 43.90\%            & 32.32\%          & 65.97\%            & 52.22\%          \\
\rowcolor{PaleGreen2} \cellcolor{White} & \cellcolor{White} & Rel. Perf. & 22.56\%     & -0.33\%   & 25.57\%            & -0.49\%          & 21.69\%            & -0.42\%          \\
Accessories   & Heavy\_Makeup            & Positive   & 6.25\%      & 6.46\%    & 35.96\%            & 32.31\%          & 55.91\%            & 52.17\%          \\
              &                          & Negative   & 7.08\%      & 6.49\%    & 34.76\%            & 32.52\%          & 54.97\%            & 52.48\%          \\
\rowcolor{PaleGreen2} \cellcolor{White} & \cellcolor{White} & Rel. Perf. & 11.70\%     & 0.46\%    & -3.44\%            & 0.62\%           & -1.71\%            & 0.59\%           \\
              & Wearing\_Hat             & Positive   & 9.01\%      & 6.24\%    & 55.58\%            & 31.23\%          & 77.17\%            & 50.65\%          \\
              &                          & Negative   & 6.05\%      & 6.49\%    & 30.40\%            & 32.54\%          & 49.86\%            & 52.55\%          \\
\rowcolor{PaleGreen2} \cellcolor{White} & \cellcolor{White} & Rel. Perf. & -48.74\%    & 3.78\%    & -82.84\%           & 4.03\%           & -54.77\%           & 3.60\%           \\
              & Wearing\_Earrings        & Positive   & 7.54\%      & 6.46\%    & 41.92\%            & 32.35\%          & 61.83\%            & 52.25\%          \\
              &                          & Negative   & 6.78\%      & 6.48\%    & 33.84\%            & 32.49\%          & 54.34\%            & 52.45\%          \\
\rowcolor{PaleGreen2} \cellcolor{White} & \cellcolor{White} & Rel. Perf. & -11.15\%    & 0.25\%    & -23.89\%           & 0.43\%           & -13.79\%           & 0.37\%           \\
              & Wearing\_Necktie         & Positive   & 3.99\%      & 6.36\%    & 19.72\%            & 31.65\%          & 37.81\%            & 51.23\%          \\
              &                          & Negative   & 7.53\%      & 6.48\%    & 41.03\%            & 32.52\%          & 62.50\%            & 52.47\%          \\
\rowcolor{PaleGreen2} \cellcolor{White} & \cellcolor{White} & Rel. Perf. & 47.05\%     & 1.88\%    & 51.93\%            & 2.65\%           & 39.51\%            & 2.37\%           \\
              & Wearing\_Lipstick        & Positive   & 6.74\%      & 6.46\%    & 38.36\%            & 32.37\%          & 58.49\%            & 52.29\%          \\
              &                          & Negative   & 7.01\%      & 6.49\%    & 34.54\%            & 32.51\%          & 54.78\%            & 52.49\%          \\
\rowcolor{PaleGreen2} \cellcolor{White} & \cellcolor{White} & Rel. Perf. & 3.91\%      & 0.39\%    & -11.05\%           & 0.44\%           & -6.78\%            & 0.39\%           \\
              & No\_Eyewear              & Positive   & 5.77\%      & 6.48\%    & 29.39\%            & 32.53\%          & 48.75\%            & 52.51\%          \\
              &                          & Negative   & 6.64\%      & 6.11\%    & 37.21\%            & 30.64\%          & 63.01\%            & 49.90\%          \\
\rowcolor{PaleGreen2} \cellcolor{White} & \cellcolor{White} & Rel. Perf. & 13.10\%     & -6.09\%   & 21.03\%            & -6.16\%          & 22.63\%            & -5.24\%          \\
              & Eyeglasses               & Positive   & 7.79\%      & 6.33\%    & 43.15\%            & 31.57\%          & 65.99\%            & 51.15\%          \\
              &                          & Negative   & 5.70\%      & 6.49\%    & 29.16\%            & 32.54\%          & 48.78\%            & 52.52\%          \\
\rowcolor{PaleGreen2} \cellcolor{White} & \cellcolor{White} & Rel. Perf. & -36.65\%    & 2.51\%    & -47.99\%           & 3.00\%           & -35.27\%           & 2.61\%           \\
Other         & Attractive               & Positive   & 6.27\%      & 6.45\%    & 36.28\%            & 32.31\%          & 56.11\%            & 52.10\%          \\
              &                          & Negative   & 7.05\%      & 6.49\%    & 34.77\%            & 32.51\%          & 54.96\%            & 52.50\%          \\
\rowcolor{PaleGreen2} \cellcolor{White} & \cellcolor{White} & Rel. Perf. & 11.16\%     & 0.51\%    & -4.35\%            & 0.61\%           & -2.09\%            & 0.77\%     \\
              \Xhline{2\arrayrulewidth}   
\end{tabular}
\end{table*}

\begin{table*}[]
\renewcommand{\arraystretch}{1.0}
\centering
\caption{ArcFace - Part 1/2. Face recognition performance based on several attributes.}
\label{tab:AttributePerformanceArcFace1}
\begin{tabular}{lllrrrrrr}
\Xhline{2\arrayrulewidth}
Category              & Attribute                         & Class  & \multicolumn{2}{c}{EER} & \multicolumn{2}{c}{FNMR@FMR=$10^{-3}$} & \multicolumn{2}{c}{FNMR@FMR=$10^{-4}$} \\
\cmidrule(rl){4-5} \cmidrule(rl){6-7} \cmidrule(rl){8-9}
              &                          &            & \multicolumn{1}{c}{Real}        & \multicolumn{1}{c}{Control}   & \multicolumn{1}{c}{Real}               & \multicolumn{1}{c}{Control}          & \multicolumn{1}{c}{Real}               & \multicolumn{1}{c}{Control}          \\
              \hline
Demographics  & Male                     & Positive   & 3.98\%       & 3.98\%       & 7.07\%            & 7.22\%            & 9.71\%            & 10.17\%           \\
              &                          & Negative   & 3.82\%       & 3.96\%       & 7.99\%            & 7.20\%            & 12.33\%           & 10.13\%           \\
\rowcolor{PaleGreen2} \cellcolor{White} & \cellcolor{White} & Rel. Perf. & -4.35\%      & -0.38\%      & 11.54\%           & -0.38\%           & 21.24\%           & -0.37\%           \\
              & Young                    & Positive   & 3.74\%       & 3.97\%       & 7.30\%            & 7.20\%            & 11.08\%           & 10.14\%           \\
              &                          & Negative   & 3.70\%       & 3.95\%       & 6.32\%            & 7.17\%            & 8.52\%            & 10.11\%           \\
\rowcolor{PaleGreen2} \cellcolor{White} & \cellcolor{White} & Rel. Perf. & -0.86\%      & -0.46\%      & -15.42\%          & -0.46\%           & -30.08\%          & -0.28\%           \\
              & Middle\_Aged             & Positive   & 3.01\%       & 3.81\%       & 5.05\%            & 6.93\%            & 6.93\%            & 9.80\%            \\
              &                          & Negative   & 4.07\%       & 3.98\%       & 7.79\%            & 7.22\%            & 11.36\%           & 10.17\%           \\
\rowcolor{PaleGreen2} \cellcolor{White} & \cellcolor{White} & Rel. Perf. & 26.14\%      & 4.05\%       & 35.20\%           & 4.04\%            & 39.04\%           & 3.56\%            \\
              & Senior                   & Positive   & 2.95\%       & 3.62\%       & 4.52\%            & 6.58\%            & 6.15\%            & 9.38\%            \\
              &                          & Negative   & 4.02\%       & 3.98\%       & 7.47\%            & 7.24\%            & 10.62\%           & 10.18\%           \\
\rowcolor{PaleGreen2} \cellcolor{White} & \cellcolor{White} & Rel. Perf. & 26.60\%      & 9.02\%       & 39.44\%           & 9.09\%            & 42.13\%           & 7.87\%            \\
              & Asian                    & Positive   & 7.99\%       & 3.29\%       & 16.68\%           & 6.01\%            & 22.59\%           & 8.69\%            \\
              &                          & Negative   & 3.73\%       & 3.98\%       & 6.75\%            & 7.23\%            & 9.61\%            & 10.18\%           \\
\rowcolor{Snow2} \cellcolor{White} & \cellcolor{White} & Rel. Perf. & -114.49\%    & 17.22\%      & -147.13\%         & 16.84\%           & -134.94\%         & 14.60\%           \\
              & White                    & Positive   & 3.27\%       & 3.98\%       & 5.84\%            & 7.23\%            & 8.55\%            & 10.18\%           \\
              &                          & Negative   & 5.80\%       & 3.91\%       & 11.69\%           & 7.10\%            & 16.03\%           & 10.01\%           \\
\rowcolor{PaleGreen2} \cellcolor{White} & \cellcolor{White} & Rel. Perf. & 43.50\%      & -1.66\%      & 50.08\%           & -1.87\%           & 46.66\%           & -1.74\%           \\
              & Black                    & Positive   & 5.72\%       & 3.40\%       & 10.90\%           & 6.21\%            & 15.02\%           & 8.95\%            \\
              &                          & Negative   & 3.85\%       & 3.98\%       & 7.06\%            & 7.23\%            & 10.11\%           & 10.18\%           \\
\rowcolor{Snow2} \cellcolor{White} & \cellcolor{White} & Rel. Perf. & -48.63\%     & 14.53\%      & -54.43\%          & 14.16\%           & -48.64\%          & 12.08\%           \\
Skin          & Rosy\_Cheeks             & Positive   & 0.98\%       & 2.91\%       & 1.17\%            & 5.12\%            & 1.31\%            & 7.47\%            \\
              &                          & Negative   & 4.39\%       & 3.98\%       & 8.33\%            & 7.23\%            & 11.77\%           & 10.16\%           \\
\rowcolor{Snow2} \cellcolor{White} & \cellcolor{White} & Rel. Perf. & 77.61\%      & 26.88\%      & 85.99\%           & 29.13\%           & 88.86\%           & 26.51\%           \\
              & Shiny\_Skin              & Positive   & 3.50\%       & 3.93\%       & 6.33\%            & 7.13\%            & 9.27\%            & 10.04\%           \\
              &                          & Negative   & 4.17\%       & 3.96\%       & 8.13\%            & 7.18\%            & 11.89\%           & 10.11\%           \\
\rowcolor{PaleGreen2} \cellcolor{White} & \cellcolor{White} & Rel. Perf. & 16.14\%      & 0.61\%       & 22.13\%           & 0.72\%            & 22.04\%           & 0.73\%            \\
Hair          & Bald                     & Positive   & 2.79\%       & 3.50\%       & 4.48\%            & 6.38\%            & 6.07\%            & 9.14\%            \\
              &                          & Negative   & 4.01\%       & 3.98\%       & 7.43\%            & 7.23\%            & 10.62\%           & 10.18\%           \\
\rowcolor{Snow2} \cellcolor{White} & \cellcolor{White} & Rel. Perf. & 30.40\%      & 12.13\%      & 39.77\%           & 11.78\%           & 42.83\%           & 10.21\%           \\
              & Wavy\_Hair               & Positive   & 3.03\%       & 3.95\%       & 6.34\%            & 7.17\%            & 10.28\%           & 10.09\%           \\
              &                          & Negative   & 4.35\%       & 3.97\%       & 7.92\%            & 7.23\%            & 10.82\%           & 10.17\%           \\
\rowcolor{PaleGreen2} \cellcolor{White} & \cellcolor{White} & Rel. Perf. & 30.46\%      & 0.73\%       & 19.95\%           & 0.84\%            & 4.96\%            & 0.80\%            \\
              & Receding\_Hairline       & Positive   & 3.03\%       & 3.92\%       & 4.68\%            & 7.12\%            & 6.13\%            & 10.04\%           \\
              &                          & Negative   & 4.10\%       & 3.97\%       & 8.20\%            & 7.22\%            & 12.25\%           & 10.15\%           \\
\rowcolor{PaleGreen2} \cellcolor{White} & \cellcolor{White} & Rel. Perf. & 26.21\%      & 1.18\%       & 42.90\%           & 1.28\%            & 49.98\%           & 1.17\%            \\
              & Bangs                    & Positive   & 4.03\%       & 3.80\%       & 8.79\%            & 6.91\%            & 13.94\%           & 9.78\%            \\
              &                          & Negative   & 3.83\%       & 3.98\%       & 6.77\%            & 7.23\%            & 9.42\%            & 10.17\%           \\
\rowcolor{PaleGreen2} \cellcolor{White} & \cellcolor{White} & Rel. Perf. & -5.11\%      & 4.43\%       & -29.80\%          & 4.44\%            & -47.96\%          & 3.89\%            \\
              & Sideburns                & Positive   & 3.72\%       & 3.97\%       & 6.51\%            & 7.21\%            & 9.10\%            & 10.13\%           \\
              &                          & Negative   & 3.98\%       & 3.97\%       & 7.62\%            & 7.22\%            & 11.10\%           & 10.16\%           \\
\rowcolor{PaleGreen2} \cellcolor{White} & \cellcolor{White} & Rel. Perf. & 6.58\%       & 0.08\%       & 14.56\%           & 0.12\%            & 18.07\%           & 0.30\%            \\
              & Black\_Hair              & Positive   & 5.12\%       & 3.92\%       & 9.85\%            & 7.11\%            & 13.47\%           & 10.01\%           \\
              &                          & Negative   & 3.48\%       & 3.97\%       & 6.36\%            & 7.21\%            & 9.28\%            & 10.15\%           \\
\rowcolor{PaleGreen2} \cellcolor{White} & \cellcolor{White} & Rel. Perf. & -47.25\%     & 1.28\%       & -54.86\%          & 1.46\%            & -45.17\%          & 1.38\%            \\
              & Blond\_Hair              & Positive   & 3.09\%       & 3.81\%       & 7.38\%            & 6.92\%            & 12.43\%           & 9.76\%            \\
              &                          & Negative   & 4.09\%       & 3.98\%       & 7.34\%            & 7.23\%            & 10.16\%           & 10.18\%           \\
\rowcolor{PaleGreen2} \cellcolor{White} & \cellcolor{White} & Rel. Perf. & 24.53\%      & 4.22\%       & -0.57\%           & 4.25\%            & -22.38\%          & 4.07\%            \\
              & Brown\_Hair              & Positive   & 3.24\%       & 3.96\%       & 6.46\%            & 7.18\%            & 10.26\%           & 10.10\%           \\
              &                          & Negative   & 4.12\%       & 3.97\%       & 7.59\%            & 7.20\%            & 10.59\%           & 10.14\%           \\
\rowcolor{PaleGreen2} \cellcolor{White} & \cellcolor{White} & Rel. Perf. & 21.36\%      & 0.35\%       & 14.93\%           & 0.26\%            & 3.11\%            & 0.36\%            \\
              & Gray\_Hair               & Positive   & 2.68\%       & 3.76\%       & 4.01\%            & 6.82\%            & 5.40\%            & 9.67\%            \\
              &                          & Negative   & 4.07\%       & 3.98\%       & 7.57\%            & 7.23\%            & 10.77\%           & 10.17\%           \\
\rowcolor{PaleGreen2} \cellcolor{White} & \cellcolor{White} & Rel. Perf. & 34.09\%      & 5.58\%       & 47.01\%           & 5.70\%            & 49.87\%           & 4.97\%            \\
Beard         & No\_Beard                & Positive   & 4.13\%       & 3.98\%       & 8.10\%            & 7.23\%            & 11.93\%           & 10.18\%           \\
              &                          & Negative   & 3.31\%       & 3.89\%       & 5.61\%            & 7.06\%            & 7.90\%            & 9.95\%            \\
\rowcolor{PaleGreen2} \cellcolor{White} & \cellcolor{White} & Rel. Perf. & -25.05\%     & -2.23\%      & -44.32\%          & -2.49\%           & -50.91\%          & -2.28\%           \\
              & Mustache                 & Positive   & 4.89\%       & 2.63\%       & 9.62\%            & 4.61\%            & 13.54\%           & 6.68\%            \\
              &                          & Negative   & 4.06\%       & 3.98\%       & 7.62\%            & 7.23\%            & 10.97\%           & 10.17\%           \\
\rowcolor{Snow2} \cellcolor{White} & \cellcolor{White} & Rel. Perf. & -20.46\%     & 33.85\%      & -26.25\%          & 36.24\%           & -23.41\%          & 34.31\%           \\
              & 5\_o\_Clock\_Shadow      & Positive   & 2.96\%       & 3.90\%       & 4.94\%            & 7.06\%            & 7.08\%            & 9.96\%            \\
              &                          & Negative   & 4.24\%       & 3.96\%       & 8.55\%            & 7.20\%            & 12.68\%           & 10.14\%           \\
\rowcolor{PaleGreen2} \cellcolor{White} & \cellcolor{White} & Rel. Perf. & 30.18\%      & 1.74\%       & 42.23\%           & 1.97\%            & 44.16\%           & 1.75\%            \\
              & Goatee                   & Positive   & 1.18\%       & 2.46\%       & 1.68\%            & 4.00\%            & 2.67\%            & 5.89\%            \\
              &                          & Negative   & 4.08\%       & 3.98\%       & 7.67\%            & 7.23\%            & 11.03\%           & 10.18\%           \\
\rowcolor{Snow2} \cellcolor{White} & \cellcolor{White} & Rel. Perf. & 71.16\%      & 38.16\%      & 78.13\%           & 44.74\%           & 75.83\%           & 42.12\%           \\
\Xhline{2\arrayrulewidth}     
\end{tabular}
\end{table*}

\begin{table*}[]
\renewcommand{\arraystretch}{1.0}
\centering
\caption{ArcFace - Part 2/2. Face recognition performance based on several attributes.}
\label{tab:AttributePerformanceArcFace2}
\begin{tabular}{lllrrrrrr}
\Xhline{2\arrayrulewidth}
Category              & Attribute                         & Class  & \multicolumn{2}{c}{EER} & \multicolumn{2}{c}{FNMR@FMR=$10^{-3}$} & \multicolumn{2}{c}{FNMR@FMR=$10^{-4}$} \\
\cmidrule(rl){4-5} \cmidrule(rl){6-7} \cmidrule(rl){8-9}
              &                          &            & \multicolumn{1}{c}{Real}        & \multicolumn{1}{c}{Control}   & \multicolumn{1}{c}{Real}               & \multicolumn{1}{c}{Control}          & \multicolumn{1}{c}{Real}               & \multicolumn{1}{c}{Control}          \\
              \hline
Face Geometry & Oval\_Face               & Positive   & 2.73\%       & 3.90\%       & 5.69\%            & 7.07\%            & 9.65\%            & 9.97\%            \\
              &                          & Negative   & 5.40\%       & 3.96\%       & 11.10\%           & 7.19\%            & 15.61\%           & 10.12\%           \\
\rowcolor{PaleGreen2} \cellcolor{White} & \cellcolor{White} & Rel. Perf. & 49.55\%      & 1.59\%       & 48.72\%           & 1.67\%            & 38.22\%           & 1.39\%            \\
              & Square\_Face             & Positive   & 3.73\%       & 3.97\%       & 6.37\%            & 7.22\%            & 8.68\%            & 10.16\%           \\
              &                          & Negative   & 4.13\%       & 3.97\%       & 8.61\%            & 7.21\%            & 13.02\%           & 10.14\%           \\
\rowcolor{PaleGreen2} \cellcolor{White} & \cellcolor{White} & Rel. Perf. & 9.65\%       & 0.03\%       & 25.96\%           & -0.10\%           & 33.37\%           & -0.15\%           \\
              & Round\_Face              & Positive   & 7.04\%       & 2.30\%       & 22.68\%           & 3.89\%            & 35.87\%           & 5.53\%            \\
              &                          & Negative   & 3.17\%       & 3.98\%       & 5.30\%            & 7.22\%            & 7.43\%            & 10.16\%           \\
\rowcolor{Snow2} \cellcolor{White} & \cellcolor{White} & Rel. Perf. & -122.46\%    & 42.18\%      & -328.09\%         & 46.22\%           & -383.05\%         & 45.63\%           \\
              & Double\_Chin             & Positive   & 3.34\%       & 3.93\%       & 5.32\%            & 7.12\%            & 7.00\%            & 10.04\%           \\
              &                          & Negative   & 4.08\%       & 3.98\%       & 7.84\%            & 7.23\%            & 11.50\%           & 10.17\%           \\
\rowcolor{PaleGreen2} \cellcolor{White} & \cellcolor{White} & Rel. Perf. & 18.22\%      & 1.23\%       & 32.24\%           & 1.43\%            & 39.15\%           & 1.29\%            \\
              & High\_Cheekbones         & Positive   & 3.34\%       & 3.95\%       & 5.96\%            & 7.17\%            & 8.63\%            & 10.10\%           \\
              &                          & Negative   & 4.28\%       & 3.97\%       & 8.60\%            & 7.20\%            & 12.70\%           & 10.13\%           \\
\rowcolor{PaleGreen2} \cellcolor{White} & \cellcolor{White} & Rel. Perf. & 21.87\%      & 0.48\%       & 30.76\%           & 0.42\%            & 32.08\%           & 0.34\%            \\
              & Chubby                   & Positive   & 3.70\%       & 3.87\%       & 6.11\%            & 7.01\%            & 7.86\%            & 9.90\%            \\
              &                          & Negative   & 3.90\%       & 3.97\%       & 7.37\%            & 7.22\%            & 10.79\%           & 10.17\%           \\
\rowcolor{PaleGreen2} \cellcolor{White} & \cellcolor{White} & Rel. Perf. & 5.18\%       & 2.62\%       & 17.14\%           & 2.85\%            & 27.15\%           & 2.58\%            \\
              & Obstructed\_Forehead     & Positive   & 5.48\%       & 3.51\%       & 13.03\%           & 6.39\%            & 20.40\%           & 9.17\%            \\
              &                          & Negative   & 3.52\%       & 3.97\%       & 6.10\%            & 7.21\%            & 8.56\%            & 10.15\%           \\
\rowcolor{Snow2} \cellcolor{White} & \cellcolor{White} & Rel. Perf. & -55.61\%     & 11.62\%      & -113.74\%         & 11.37\%           & -138.28\%         & 9.66\%            \\
              & Fully\_Visible\_Forehead & Positive   & 3.30\%       & 3.97\%       & 5.47\%            & 7.21\%            & 7.49\%            & 10.15\%           \\
              &                          & Negative   & 4.64\%       & 3.95\%       & 9.98\%            & 7.16\%            & 15.06\%           & 10.06\%           \\
\rowcolor{PaleGreen2} \cellcolor{White} & \cellcolor{White} & Rel. Perf. & 28.85\%      & -0.46\%      & 45.15\%           & -0.70\%           & 50.30\%           & -0.86\%           \\
Periocular    & Brown\_Eyes              & Positive   & 4.69\%       & 3.97\%       & 9.13\%            & 7.21\%            & 12.88\%           & 10.14\%           \\
              &                          & Negative   & 2.63\%       & 3.85\%       & 5.36\%            & 6.98\%            & 8.73\%            & 9.85\%            \\
\rowcolor{PaleGreen2} \cellcolor{White} & \cellcolor{White} & Rel. Perf. & -78.48\%     & -2.96\%      & -70.17\%          & -3.28\%           & -47.51\%          & -2.92\%           \\
              & Bags\_Under\_Eyes        & Positive   & 3.78\%       & 3.96\%       & 6.37\%            & 7.19\%            & 8.48\%            & 10.11\%           \\
              &                          & Negative   & 3.87\%       & 3.96\%       & 8.17\%            & 7.19\%            & 12.63\%           & 10.12\%           \\
\rowcolor{PaleGreen2} \cellcolor{White} & \cellcolor{White} & Rel. Perf. & 2.20\%       & 0.13\%       & 22.05\%           & 0.01\%            & 32.84\%           & 0.10\%            \\
              & Bushy\_Eyebrows          & Positive   & 3.51\%       & 3.96\%       & 6.05\%            & 7.19\%            & 8.28\%            & 10.11\%           \\
              &                          & Negative   & 4.00\%       & 3.97\%       & 7.81\%            & 7.22\%            & 11.58\%           & 10.17\%           \\
\rowcolor{PaleGreen2} \cellcolor{White} & \cellcolor{White} & Rel. Perf. & 12.35\%      & 0.20\%       & 22.54\%           & 0.54\%            & 28.46\%           & 0.60\%            \\
              & Arched\_Eyebrows         & Positive   & 3.21\%       & 3.94\%       & 6.08\%            & 7.15\%            & 9.29\%            & 10.05\%           \\
              &                          & Negative   & 4.42\%       & 3.97\%       & 8.38\%            & 7.23\%            & 11.79\%           & 10.17\%           \\
\rowcolor{PaleGreen2} \cellcolor{White} & \cellcolor{White} & Rel. Perf. & 27.52\%      & 0.81\%       & 27.46\%           & 1.05\%            & 21.20\%           & 1.14\%            \\
Mouth         & Mouth\_Closed            & Positive   & 3.06\%       & 3.37\%       & 5.40\%            & 6.13\%            & 7.70\%            & 8.85\%            \\
              &                          & Negative   & 3.88\%       & 3.90\%       & 7.79\%            & 7.08\%            & 11.93\%           & 9.99\%            \\
\rowcolor{Snow2} \cellcolor{White} & \cellcolor{White} & Rel. Perf. & 21.21\%      & 13.69\%      & 30.62\%           & 13.38\%           & 35.48\%           & 11.39\%           \\
              & Smiling                  & Positive   & 3.35\%       & 3.94\%       & 5.93\%            & 7.14\%            & 8.48\%            & 10.05\%           \\
              &                          & Negative   & 4.62\%       & 3.96\%       & 9.65\%            & 7.19\%            & 14.57\%           & 10.12\%           \\
\rowcolor{PaleGreen2} \cellcolor{White} & \cellcolor{White} & Rel. Perf. & 27.32\%      & 0.56\%       & 38.59\%           & 0.71\%            & 41.81\%           & 0.65\%            \\
              & Big\_Lips                & Positive   & 4.15\%       & 3.94\%       & 8.12\%            & 7.17\%            & 11.91\%           & 10.10\%           \\
              &                          & Negative   & 3.88\%       & 3.97\%       & 7.00\%            & 7.22\%            & 9.84\%            & 10.16\%           \\
\rowcolor{PaleGreen2} \cellcolor{White} & \cellcolor{White} & Rel. Perf. & -6.93\%      & 0.78\%       & -16.08\%          & 0.75\%            & -21.01\%          & 0.63\%            \\
Nose          & Big\_Nose                & Positive   & 4.39\%       & 3.90\%       & 7.89\%            & 7.07\%            & 10.48\%           & 9.95\%            \\
              &                          & Negative   & 3.90\%       & 3.95\%       & 8.62\%            & 7.18\%            & 13.58\%           & 10.10\%           \\
\rowcolor{PaleGreen2} \cellcolor{White} & \cellcolor{White} & Rel. Perf. & -12.67\%     & 1.49\%       & 8.52\%            & 1.51\%            & 22.80\%           & 1.46\%            \\
              & Pointy\_Nose             & Positive   & 3.15\%       & 3.97\%       & 5.84\%            & 7.22\%            & 8.86\%            & 10.16\%           \\
              &                          & Negative   & 5.28\%       & 3.96\%       & 10.46\%           & 7.18\%            & 14.62\%           & 10.11\%           \\
\rowcolor{PaleGreen2} \cellcolor{White} & \cellcolor{White} & Rel. Perf. & 40.44\%      & -0.43\%      & 44.19\%           & -0.63\%           & 39.44\%           & -0.49\%           \\
Accessories   & Heavy\_Makeup            & Positive   & 3.08\%       & 3.96\%       & 5.79\%            & 7.20\%            & 9.00\%            & 10.13\%           \\
              &                          & Negative   & 4.32\%       & 3.97\%       & 8.02\%            & 7.22\%            & 11.14\%           & 10.16\%           \\
\rowcolor{PaleGreen2} \cellcolor{White} & \cellcolor{White} & Rel. Perf. & 28.75\%      & 0.18\%       & 27.75\%           & 0.24\%            & 19.27\%           & 0.32\%            \\
              & Wearing\_Hat             & Positive   & 5.51\%       & 3.66\%       & 12.28\%           & 6.62\%            & 18.45\%           & 9.44\%            \\
              &                          & Negative   & 3.71\%       & 3.98\%       & 6.53\%            & 7.23\%            & 9.14\%            & 10.18\%           \\
\rowcolor{PaleGreen2} \cellcolor{White} & \cellcolor{White} & Rel. Perf. & -48.79\%     & 8.09\%       & -88.01\%          & 8.45\%            & -101.94\%         & 7.29\%            \\
              & Wearing\_Earrings        & Positive   & 3.25\%       & 3.95\%       & 6.64\%            & 7.17\%            & 10.59\%           & 10.10\%           \\
              &                          & Negative   & 4.08\%       & 3.98\%       & 7.33\%            & 7.23\%            & 10.10\%           & 10.17\%           \\
\rowcolor{PaleGreen2} \cellcolor{White} & \cellcolor{White} & Rel. Perf. & 20.23\%      & 0.83\%       & 9.44\%            & 0.80\%            & -4.92\%           & 0.78\%            \\
              & Wearing\_Necktie         & Positive   & 2.72\%       & 3.82\%       & 3.84\%            & 6.92\%            & 4.72\%            & 9.79\%            \\
              &                          & Negative   & 4.25\%       & 3.97\%       & 8.52\%            & 7.22\%            & 12.68\%           & 10.16\%           \\
\rowcolor{PaleGreen2} \cellcolor{White} & \cellcolor{White} & Rel. Perf. & 35.95\%      & 4.00\%       & 54.94\%           & 4.24\%            & 62.77\%           & 3.73\%            \\
              & Wearing\_Lipstick        & Positive   & 3.28\%       & 3.96\%       & 6.38\%            & 7.19\%            & 9.93\%            & 10.11\%           \\
              &                          & Negative   & 4.27\%       & 3.98\%       & 7.85\%            & 7.23\%            & 10.83\%           & 10.18\%           \\
\rowcolor{PaleGreen2} \cellcolor{White} & \cellcolor{White} & Rel. Perf. & 23.21\%      & 0.40\%       & 18.74\%           & 0.57\%            & 8.25\%            & 0.65\%            \\
              & No\_Eyewear              & Positive   & 3.64\%       & 3.98\%       & 6.39\%            & 7.23\%            & 8.92\%            & 10.18\%           \\
              &                          & Negative   & 3.86\%       & 3.50\%       & 6.62\%            & 6.35\%            & 8.99\%            & 9.12\%            \\
\rowcolor{Snow2} \cellcolor{White} & \cellcolor{White} & Rel. Perf. & 5.75\%       & -13.57\%     & 3.42\%            & -13.84\%          & 0.82\%            & -11.60\%          \\
              & Eyeglasses               & Positive   & 4.60\%       & 3.79\%       & 9.13\%            & 6.88\%            & 13.03\%           & 9.75\%            \\
              &                          & Negative   & 3.68\%       & 3.98\%       & 6.45\%            & 7.23\%            & 8.99\%            & 10.18\%           \\
\rowcolor{PaleGreen2} \cellcolor{White} & \cellcolor{White} & Rel. Perf. & -25.08\%     & 4.88\%       & -41.53\%          & 4.89\%            & -44.86\%          & 4.24\%            \\
Other         & Attractive               & Positive   & 2.95\%       & 3.96\%       & 5.49\%            & 7.19\%            & 8.60\%            & 10.10\%           \\
              &                          & Negative   & 4.30\%       & 3.97\%       & 8.02\%            & 7.22\%            & 11.14\%           & 10.17\%           \\
\rowcolor{PaleGreen2} \cellcolor{White} & \cellcolor{White} & Rel. Perf. & 31.44\%      & 0.20\%       & 31.57\%           & 0.47\%            & 22.82\%           & 0.65\%    \\
              \Xhline{2\arrayrulewidth}   
\end{tabular}
\end{table*}

\begin{figure*}
\centering
\subfloat[FaceNet \label{fig:SummaryFaceNet}]{%
       \includegraphics[width=0.99\textwidth]{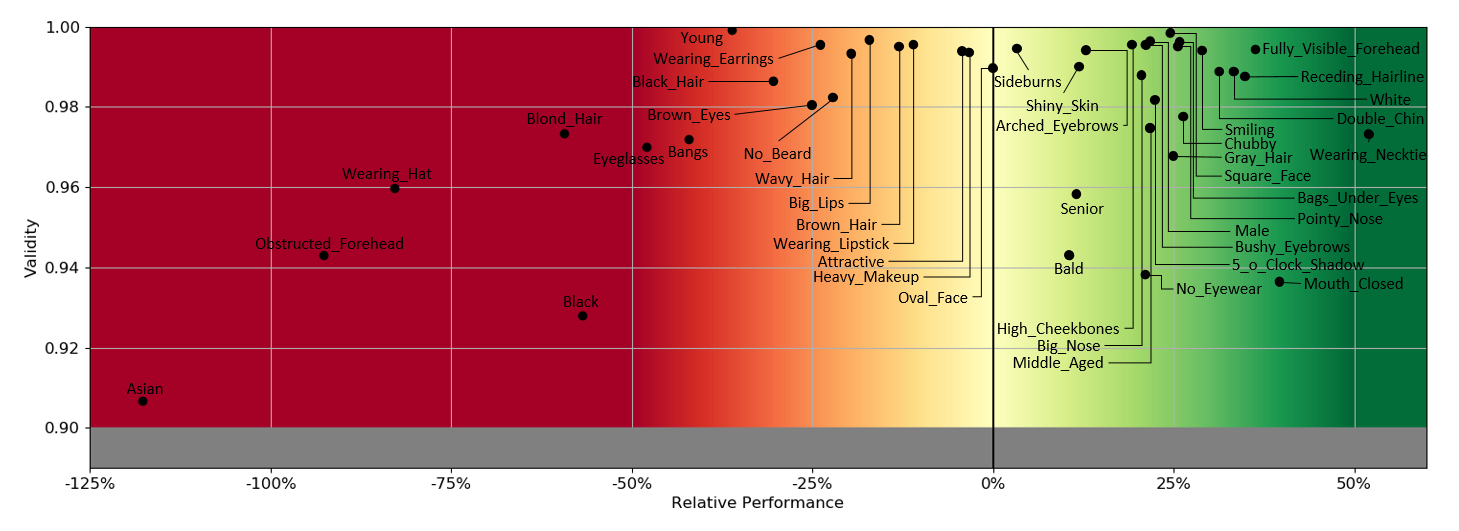}}
       
\subfloat[ArcFace \label{fig:SummaryArcFace}]{%
       \includegraphics[width=0.99\textwidth]{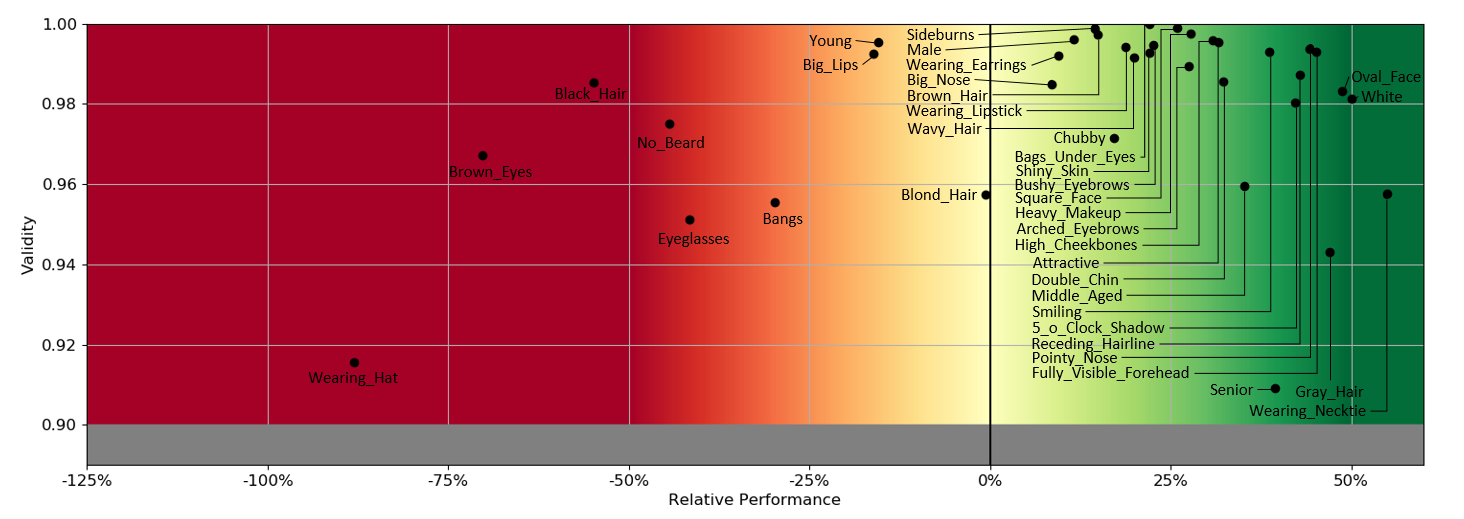}}
\caption{Visual summary on the differential outcome affected by each attribute. Figure \ref{fig:SummaryFaceNet} visualizes the results for FaceNet, while Figure \ref{fig:SummaryArcFace} visualizes the results for ArcFace. The relative performance is based on the recognition performance on the positively-labelled data versus the performance of negatively-labelled data. The validity is based on the performance differences of the control groups. Validity values below 0.9 (more than 10\% performance differences between the control groups) are considered as \textit{not valid} (grey area) and are not shown in this figure. The red areas indicate an attribute-related bias that leads to a degraded face recognition performance for faces with the specific attribute. Green areas indicate that faces possessing a specific attribute enhances the recognition performance. It can be observed that the majority of the investigated attributes strongly affects the recognition performance.}
\label{fig:Summary}
\end{figure*}

\subsection{Performance Analysis}
To provide an overview of the findings, Figure \ref{fig:Summary} shows the relative performance differences on FaceNet and ArcFace features based on the investigated attributes.
The shown relative performance is based on the FMR at $10^{-3}$ FNMR as recommended by the  European Boarder Guard Agency Frontex \cite{FrontexBestPractice}.
The validity describes the performance difference between the positive and negative attribute-related control groups as shown in Equation \ref{eq:validity}.
An attribute performance with a validity of less then $90\%$ is considered as \textit{not valid} (grey area) since the unbalanced data annotations might affect the reported performance.
The red area indicates that the recognition performance of the positive attribute class is significantly weaker than the performance of the negative class.
In contrast, the green area indicates a significant improvement of recognition performance of the positive attribute class over the negative class.
If an attribute has only a minor effect on the recognition performance, the relative performance is close to 0\% (yellow area).

\subsubsection{FaceNet vs ArcFace}
The main difference between FaceNet and ArcFace are the underlying training-principles.
FaceNet uses triplet-loss learning \cite{DBLP:journals/corr/SchroffKP15} that aims solely at minimizing the intra-class variations while maximizing the inter-class variations.
In contrast, ArcFace introduces an angular large-margin principle \cite{Deng_2019_CVPR} that additionally aims at enhancing the robustness of recognition model. 
The utilized training principle together with the used network structure and the training data determines the recognition behaviour.
This includes the effect of differential outcome appearing when certain attributes of the face are present.
Since the used FaceNet and ArcFace models share the same network structure and training data, the observed differential outcome might arise from the training principles.

\subsubsection{The Effect of Attributes on Recognition}
It turns out that the majority of the investigated attributes strongly affect the recognition performance of both, FaceNet and ArcFace.
For FaceNet, many faces that are perceived as \textit{Attractive} or make use of \textit{Heavy Makeup} do not show to alter the recognition performance unlike previously reported in \cite{DBLP:journals/access/RathgebDB19}.
The same goes for \textit{Oval Faces} and faces with \textit{Sideburns}.
For ArcFace, \textit{Blond Hair}, \textit{Big Nose}, \textit{Big Lips}, \textit{Wearing Earrings}, and \textit{Young} faces show only a minor effect on the recognition performance.
For both recognition models, the majority of the investigated attributes strongly affect the recognition performance.
Some of the observations might be explainable.

\begin{itemize}
\item \textbf{Demographics:} Recent works \cite{DBLP:conf/btas/HowardSV19, DBLP:conf/cvpr/RobinsonLHQ0T20, FRVT2019, DBLP:journals/corr/abs-2007-06570} extensively discussed the impact of demographic attributes on face recognition. Our results support the findings from previous works. We observe an improved recognition performance for the attributes \textit{Middle Aged}, \textit{Senior}, \textit{White}, and \textit{Male}. Contrarily, a degraded recognition performance is observed for \textit{Young}, \textit{Asian}, \textit{Black}, and \textit{Female} faces.
For FaceNet, the observed differential outcome are stronger than for ArcFace.
Moreover, we could not show that \textit{Asian} or \textit{Black} faces perform weaker than \textit{White} faces on ArcFace, since the data unbalance lead to a low validity for our results.

\item \textbf{Visibility-related attributes:} We observe that attributes that indicate a fully visible face lead to an improved face recognition performance. This includes the attributes \textit{Fully Visible Forehead}, \textit{Receding Hairline}, \textit{No Eyewear}, and \textit{Bald}.
In contrast, attributes that might lead to small partial occlusions of the face lead to significantly degraded recognition performances \cite{face-regions}.
For FaceNet, this includes faces with an \textit{Obstructed Forehead}, \textit{Bangs}, and \textit{Wavy Hair}.
For ArcFace, this includes samples with \textit{Eyeglasses} or \textit{Bangs}.

\item \textbf{Temporary attributes:} For faces with temporary attributes, such as for accessories, a degraded face recognition performance can be observed.
This includes \textit{Wearing Hat}, \textit{Wearing Earrings}, \textit{Wearing Lipstick}, and \textit{Eyeglasses}.
Beside a partial-occlusion of small parts of the face, these attributes are non-permanent and can quickly change the appearance of the face.

\item \textbf{Anomalous characteristics:} It turns out that conspicuous characteristics that is only possessed by a small proportion of the population lead to strongly enhanced recognition performances. 
This includes \textit{Arched Eyebrows}, \textit{Big Nose}, \textit{Pointy Nose}, \textit{Bushy Eyebrows}, \textit{Double Chin}, and \textit{High Cheekbones} \cite{maadface}.

\item \textbf{Facial expressions:} Faces that are \textit{Smiling} or that have their \textit{Mouth Closed} perform above average for face recognition.
However, faces with non-neutral expressions lead to degraded face recognition performances.
This bias might come from the data utilized for training that usually contains neutral or smiling faces and was discussed in more details by previous works \cite{DBLP:journals/pami/ChangBF06, 10.1117/12.604171}.

\end{itemize}

While these attribute-dependent differential outcome might be explainable, the reason for the impact of other attributes on recognition is currently unclear.

\begin{itemize}
\item \textbf{Colors:} The results demonstrate strong differential outcome based on the user's hair- and eyecolor.
For FaceNet, faces with \textit{Blond Hair}, \textit{Black Hair}, and \textit{Brown Hair} show strongly degraded recognition performances.
In contrast, faces with \textit{Gray Hair} lead to an improved recognition.
For ArcFace, \textit{Gray Hair} also strongly improves the recognition performance while \textit{Black Hair} decreases it.
The differential outcome for \textit{Blond Hair} and \textit{Brown Hair} strongly varies dependent on the used decision threshold.
For instance, for high FMRs, \textit{Blond Hair} has a positive effect on recognition, for a lower FMR (e.g. $10^{-4}$) the same attribute changes to a negative effect.
The same can be observed for eyecolors. 
Faces with \textit{Brown Eyes} perform weaker than faces from the opposite group.
The differential outcome of these attributes does not reflect the distribution of the training data and thus, might arise from a different origin.

\item \textbf{Beard:} As we discussed before, attributes that might induce a partially occluded face lead to a degraded face recognition performance.
Although, beards can cover parts of the face, the results demonstrate the faces with \textit{No Beard} perform below-average, while faces with e.g. a \textit{5 o Clock Shadow} achieve much higher recognition rates.

\item \textbf{Wearing Necktie:} Unlike other accessories, \textit{Wearing Necktie} improved the face recognition performance drastically.
We assume that this might results from a data collection bias induced by the correlation with hidden factors, such as environment.
Persons who present themselves in public (e.g. celebrities) might often wear a necktie and thus, photos are often taken with frontal poses and full lightning.
However, the high validity and the strong differential outcome makes it hard to argue in this direction.

\item \textbf{Antagonistic Behaviour:}
Some attributes might result in differential outcome of the opposite direction depending on the used training-principle (triplet vs angular margin loss).
For instance, faces with \textit{Wavy Hair} lead to a negative performance on FaceNet and to positive performance on ArcFace.
Also the attributes \textit{Attractiveness}, \textit{Heavy Makeup}, and \textit{Oval Faces} negatively affect the recognition performance on FaceNet, but show some strong positive impacts on the recognition performance of ArcFace.

\end{itemize}

As mentioned earlier, the resulting performance of a face recognition model is mainly determined by its loss-function, its network architecture, and the utilized training data.
Since both investigated models have the last two points in common, the observed differences in the performance might arise from the underlying training principles.
Generally, we observe that the large angular margin loss from ArcFace leads to a significantly stronger overall recognition performance compared to FaceNet.
The loss aiming to enhance the model robustness also shows a clearly visible effect on the attribute-related differential outcome.
On ArcFace, slightly less attributes negatively affect the recognition performance than on FaceNet.
However, the differential outcome that origins from the affected (biased) attributes are still of high impact.
A remarkable observation is the fact that the differential outcome remain relatively constant over several decision thresholds for FaceNet, while for ArcFace the differential outcome often significantly vary for different decision thresholds.
This can be observed for instance for faces with \textit{Bangs}, \textit{Blond Hair}, or a \textit{Double Chin}.

%
%


%





\subsubsection{Future challenges for face recognition}

The observations of the experiment point out some critical issues of current face recognition solutions in terms of robustness, fairness, and explainability.
\begin{itemize}
\item \textbf{Need for robustness:} Face recognition systems need to become more robust against partial occlusions (from accessories or hair) \cite{face-regions, DBLP:conf/iccv/LiuLWT15}, facial expressions (beyond neutral and smiling faces) \cite{face-emo}, and temporary attributes that might change the daily appearance of a face \cite{Tan, Nixon}. This can greatly enhance the applicability in more real-life scenarios.
\item \textbf{Need for fairness:} Face recognition systems need to enhance the user-fairness.
We observed differential outcome based on the user-demographics (demographic-bias), anomalous characteristics (such as pointy noses, bushy eyebrows, and high cheekbones), beard types, and accessories. 
This can lead to discriminative decisions \cite{Discrimination} of face recognition systems that several political regulation, such as the GDPR \cite{Voigt:2017:EGD:3152676}, try to prevent.
\item \textbf{Need for explainability:} Face recognition models need to explain themselves. Why do colors/face shapes/beards/accessories lead to differential outcome? Why can we observe an antagonistic behaviour between the two different learning principles for some attributes? In order to enhance the model transparency and to enable efficient model-debugging, future work have to elaborate on the explainability \cite{XAI, XAI-symbolic} of face recognition models.
\item \textbf{Need for comprehensive approaches and transfer learning:} The previous areas related to robustness, fairness, and explainability will significantly benefit from more comprehensive approaches that consider simultaneously all the elements and attributes in place \cite{fusion1, fusion2}, exploiting at the same time previous or general knowledge of the problem at hand \cite{transfer1, transfer2}. Most of the research so far in biometrics bias, especially around face biometrics, has been mainly oriented to studying individual elements (e.g., gender or ethnicity) not exploiting previous models or evidence. There is a need for more comprehensive approaches like the one presented here (incorporating simultaneously 47 relevant attributes) and new schemes to easily exploit the generated knowledge.
\end{itemize}

%
%
%

\vspace{-2mm}

\section{Conclusion}
The growing effect of face recognition systems on the daily life, including critical decision-making processes, shows the need of non-discriminative face recognition solutions.
Previous works focused on estimating and mitigating demographic-bias.
However, to deploy non-discriminatory face recognition systems, it is necessary to know which differential outcome appear in the presences of certain facial attributes beyond demographics.
Driven by this need, we analysed the performance differences on two popular face recognition models concerning 47 different attributes.
The experiment was conducted on the publicly available MAAD-Face database, a large-scale dataset with over 120M attribute annotations of high-quality.
To prevent misleading statements of attribute biases, we consider attribute correlations and  minimize the effect of unbalanced testing data via control group based validity values.
We investigated the effect of two different learning-principles on the differential outcome originating from facial attributes.
The results show that, besides demographics, many attributes strongly affect the recognition performance of both investigated face recognition models, FaceNet and ArcFace.
While for FaceNet the observed differential outcome originated by several attributes remain relatively constant, these differences strongly depend on the used decision threshold for ArcFace.
We provided explanations for many observed performance differences.
However, the reason for some observations remain unclear and have to be addressed by future work.
The findings of this work strongly demonstrate the need for further advances in making face recognition systems more robust, explainable, and fair. 
We hope these findings lead to the development of more robust and unbiased face recognition solutions.

\paragraph*{Acknowledgment}
This research work has been funded by the German Federal Ministry of Education and Research and the Hessian Ministry of Higher Education, Research, Science and the Arts within their joint support of the National Research Center for Applied Cybersecurity ATHENE.
This was also funded from the projects BIBECA (RTI2018-101248-B-I00 MINECO/FEDER) and PRIMA (H2020-MSCA-ITN-2019-860315).

\ifCLASSOPTIONcaptionsoff
  \newpage
\fi



%
%
%

\vspace{-2mm}

{\small
\bibliographystyle{ieee}
\bibliography{egbib}
}

\end{document}